\pdfoutput=1
\documentclass[11pt]{article}
\usepackage{times}
\usepackage{amsmath}
\usepackage{url}
\usepackage{xspace}
\usepackage{latexsym}
\usepackage{booktabs}
\usepackage{multirow}
\usepackage{caption}
\usepackage{graphicx}
\usepackage{fix-cm}
\usepackage[table]{xcolor}

\usepackage{ACL2023}
\usepackage{times}
\usepackage{latexsym}

\usepackage[T1]{fontenc}
\usepackage[utf8]{inputenc}
\usepackage{microtype}
\usepackage{inconsolata}

\usepackage{amsthm}

\usepackage{algorithm}
\usepackage[noend]{algpseudocode}

\usepackage{enumitem}

\usepackage{multirow}
\usepackage{adjustbox}
\usepackage{bbm}
\usepackage{wrapfig,lipsum}
\usepackage{titlesec}
\usepackage{booktabs,cellspace}  
\usepackage[normalem]{ulem}
\usepackage{makecell, cellspace, caption}
\usepackage{amssymb}
\usepackage{pifont}
\definecolor{lightblue}{HTML}{F2FBFF}
\definecolor{blue}{HTML}{CFF8FF}

\usepackage{microtype}

\usepackage{cleveref}

\newcommand{\stitle}[1]{\vspace{1ex} \noindent{\bf #1.}}

\newcommand{\gpttwo}{\textsc{GPT-2}\xspace}
\newcommand{\gptthreepointfive}{\textsc{GPT-3.5}\xspace}
\newcommand{\gptfour}{\textsc{GPT-4}\xspace}
\newcommand{\gptfouro}{\textsc{GPT-4o}\xspace}
\newcommand{\gptfouromini}{\textsc{GPT-4o-mini}\xspace}

\newcommand{\summarization}{\textsc{summarization}\xspace}
\newcommand{\improvement}{\textsc{improvement}\xspace}
\newcommand{\translation}{\textsc{translation}\xspace}
\newcommand{\refinement}{\textsc{refinement}\xspace}

\newcommand{\llamathree}{\textsc{Llama3}\xspace}
\newcommand{\llamathreepointone}{\textsc{Llama3.1}\xspace}
\newcommand{\qwenonepointfive}{\textsc{Qwen-1.5}\xspace}
\newcommand{\gemma}{\textsc{Gemma}\xspace}
\newcommand{\gemmatwo}{\textsc{Gemma2}\xspace}
\newcommand{\mixtral}{\textsc{Mixtral}\xspace}
\newcommand{\qwen}{\textsc{Qwen}\xspace}
\newcommand{\qwentwo}{\textsc{Qwen2}\xspace}

\newcommand{\gemmatwonineb}{\textsc{Gemma2-9B}\xspace}
\newcommand{\gemmatwosevenb}{\textsc{Gemma2-27B}\xspace}
\newcommand{\qwenseventwob}{\textsc{Qwen-1.5-72B}\xspace}
\newcommand{\qwenoneonezerob}{\textsc{Qwen-1.5-110B}\xspace}
\newcommand{\llamathreeoneseventyb}{\textsc{Llama3.1-70B}\xspace}
\newcommand{\qwentst}{\textsc{Qwen-2-72B}\xspace}
\newcommand{\mixtralone}{\textsc{Mixtral-8x22B}\xspace}
\newcommand{\gemmatts}{\textsc{Gemma2-27B}\xspace}

\newcommand{\llamaapimixtral}{\textsc{Mixtral-8x22B-instruct}\xspace}
\newcommand{\llamaapiqwentst}{\textsc{Qwen2-72B}\xspace}
\newcommand{\llamaapiqwenoooz}{\textsc{Qwen1.5-110B-Chat}\xspace}
\newcommand{\llamaapiqwenost}{\textsc{Qwen1.5-72B-Chat}\xspace}

\newcommand{\llamaapigemmatts}{\textsc{Gemma2-27B}\xspace}

\newcommand{\groqmixtral}{\textsc{Mixtral-8x7B-32768}\xspace}
\newcommand{\groqllamate}{\textsc{Llama3-8B-8192}\xspace}
\newcommand{\groqllamatoe}{\textsc{Llama-3.1-8B-instant}\xspace}
\newcommand{\groqllamatsz}{\textsc{Llama3-70B-8192}\xspace}
\newcommand{\groqllamatosz}{\textsc{Llama-3.1-70B-versatile}\xspace}
\newcommand{\groqgemmas}{\textsc{Gemma-7B-it}\xspace}
\newcommand{\groqgemmatn}{\textsc{Gemma2-9B-it}\xspace}

\newcommand{\sitr}{\textsc{SITR}\xspace}
\newcommand{\cona}{\textsc{w/o [summarization prompt]}\xspace}
\newcommand{\conb}{\textsc{w/o [translation prompt]}\xspace}
\newcommand{\conc}{\textsc{w/o both}\xspace}

\usepackage{etoolbox}
\usepackage[tikz]{bclogo}

\usepackage{subcaption}
\usepackage{adjustbox}

\usepackage{tcolorbox} 
\tcbuselibrary{skins} 
\usepackage[T1]{fontenc}

\definecolor{msftBlue}{RGB}{0,164,239}
\definecolor{msftGreen}{RGB}{127,186,0}
\definecolor{msftYello}{RGB}{255,185,0}
\definecolor{msftBlack}{RGB}{0,0,0}

\tcbset{
    userstyle/.style={
        enhanced,
        colback=white,
        colframe=black,
        colbacktitle=gray!20,
        coltitle=black,
        rounded corners,
        sharp corners=north,
        boxrule=0.5pt,
        drop shadow=black!50!white,
        attach boxed title to top left={
            xshift=-2mm,
            yshift=-2mm
        },
        boxed title style={
            rounded corners,
            size=small,
            colback=gray!20
        }
    },
    replystyleg/.style={
        enhanced,
        colback=green!15,
        colframe=black,
        colbacktitle=green!30,
        coltitle=black,
        boxrule=0.5pt,
        drop shadow=black!50!white,
        rounded corners,
        sharp corners=north,
        attach boxed title to top right={
            xshift=-2mm,
            yshift=-2mm
        },
        boxed title style={
            rounded corners,
            size=small,
            colback=green!40
        }
    },
    replystyler/.style={
        enhanced,
        colback=red!15,
        colframe=black,
        colbacktitle=red!40,
        coltitle=black,
        boxrule=0.5pt,
        drop shadow=black!50!white,
        rounded corners,
        sharp corners=north,
        attach boxed title to top right={
            xshift=-2mm,
            yshift=-2mm
        },
        boxed title style={
            rounded corners,
            size=small,
            colback=red!40
        }
    }
}

\newtcolorbox{userquery}[1][]{
    userstyle,
    title=Prompt,
    #1
}

\title{Think Carefully and Check Again! Meta-Generation Unlocking LLMs for Low-Resource Cross-Lingual Summarization}

\author{
Zhecheng Li$^\dagger$ \ \ \ \ Yiwei Wang$^\ddagger$ \ \ \ \ Bryan Hooi$^\|$ \ \ \ \ Yujun Cai$^\mathsection$ \\ 
\textbf{Naifan Cheung}$^\mathparagraph$ \ \ \ \ \textbf{Nanyun Peng}$^\ddagger$ \ \ \ \ \textbf{Kai-Wei Chang}$^\ddagger$ \\
$^\dagger$ University of California, San Diego \quad $^\ddagger$ University of California, Los Angelas \\
$^\mathsection$ Nanyang Technological University \quad $^\|$ National University of Singapore \\
$^\mathparagraph$ Tsinghua University \\
\texttt{zhl186@ucsd.edu}
}

\begin{document}
\maketitle
\begin{abstract}
Cross-lingual summarization (CLS) aims to generate a summary for the source text in a different target language. Currently, instruction-tuned large language models (LLMs) excel at various English tasks. 
However, unlike languages such as English, Chinese or Spanish, for those relatively low-resource languages with limited usage or data, recent studies have shown that LLMs' performance on CLS tasks remains unsatisfactory even with few-shot settings. This raises the question: \textit{Are LLMs capable of handling cross-lingual summarization tasks for low-resource languages?}
To resolve this question, we fully explore the potential of large language models on cross-lingual summarization task for low-resource languages through our four-step zero-shot method: \textbf{{\summarization}}, \textbf{{\improvement}}, \textbf{{\translation}} and \textbf{{\refinement}} ({\sitr}) with correspondingly designed prompts.
We test our proposed method with multiple LLMs on two well-known cross-lingual summarization datasets with various low-resource target languages. The results show that: 
i) {\gptthreepointfive} and {\gptfour} significantly and consistently outperform other baselines when using our zero-shot \sitr methods. 
ii) By employing our proposed method, we unlock the potential of LLMs, enabling them to effectively handle cross-lingual summarization tasks for relatively low-resource languages. 
\end{abstract}

\section{Introduction}
\begin{figure*}[!tb]
	\centering
	\includegraphics[width=1\linewidth]{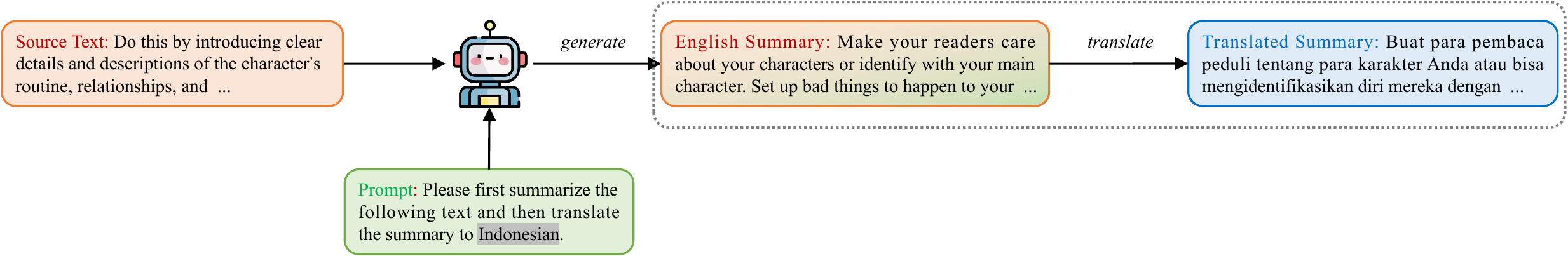}
	\caption{An example of single-step summarize-translate method for cross-lingual summarization.
    \vspace{-2mm}
	\label{fig:summarizetranslate}}
\end{figure*}

Cross-lingual summarization refers to summarizing the source text in another target language. 
Traditionally, CLS is approached through one of two methods: summarize-translate (see LLM implementation in Figure \ref{fig:summarizetranslate}) or translate-summarize \cite{leuski, orasan}. In the summarize-translate method, the text is first summarized in the source language and then translated into the target language. The translate-summarize method reverses this order. Both approaches, however, are prone to error accumulation during the two-step process, which can significantly degrade the final output quality.


With the advent of the Transformer architecture \cite{transformer}, end-to-end multilingual models like mBART \cite{mbartliu}, mBART-50 \cite{mbart50tang}, and mT5 \cite{mt5} have been developed and applied to CLS tasks. However, these models often require extensive fine-tuning, especially when applied to low-resource languages with limited pre-training data \cite{sumtra}.

In recent years, large language models (LLMs) such as \gpttwo, InstructGPT, \gptfour, and Llama \cite{gpt2, gpt3, instructgpt, gpt4, llama3} have shown significant potential for CLS tasks due to their extensive training on vast multilingual data. These models have achieved strong performance in high-resource languages like English, Chinese, and German \cite{wangzeroshot} by implementing summarize-translate method. However, their effectiveness in low-resource languages remains limited, even when using few-shot learning techniques \cite{park}. 


This limitation underscores a critical area of research that has not yet been fully explored: whether LLMs can be effectively adapted for cross-lingual summarization tasks in low-resource languages, and if so, how effective they can be. Addressing this gap is crucial for extending the benefits of LLMs to a broader range of linguistic communities, making it an important area for further investigation.

To address these challenges, we propose a four-step zero-shot approach, \textit{Summarization}, \textit{Improvement}, \textit{Translation}, and \textit{Refinement} (\sitr) — designed to unlock the full potential of LLMs for CLS tasks in low-resource languages. Our method mitigates the issues of traditional pipelines by incorporating meta-generation strategies, which allows LLMs to learn from feedback and use refiners to produce more accurate and reliable outputs \cite{metageneration}.

\begin{figure*}[!tb]
	\centering
	\includegraphics[width=1.00\linewidth]{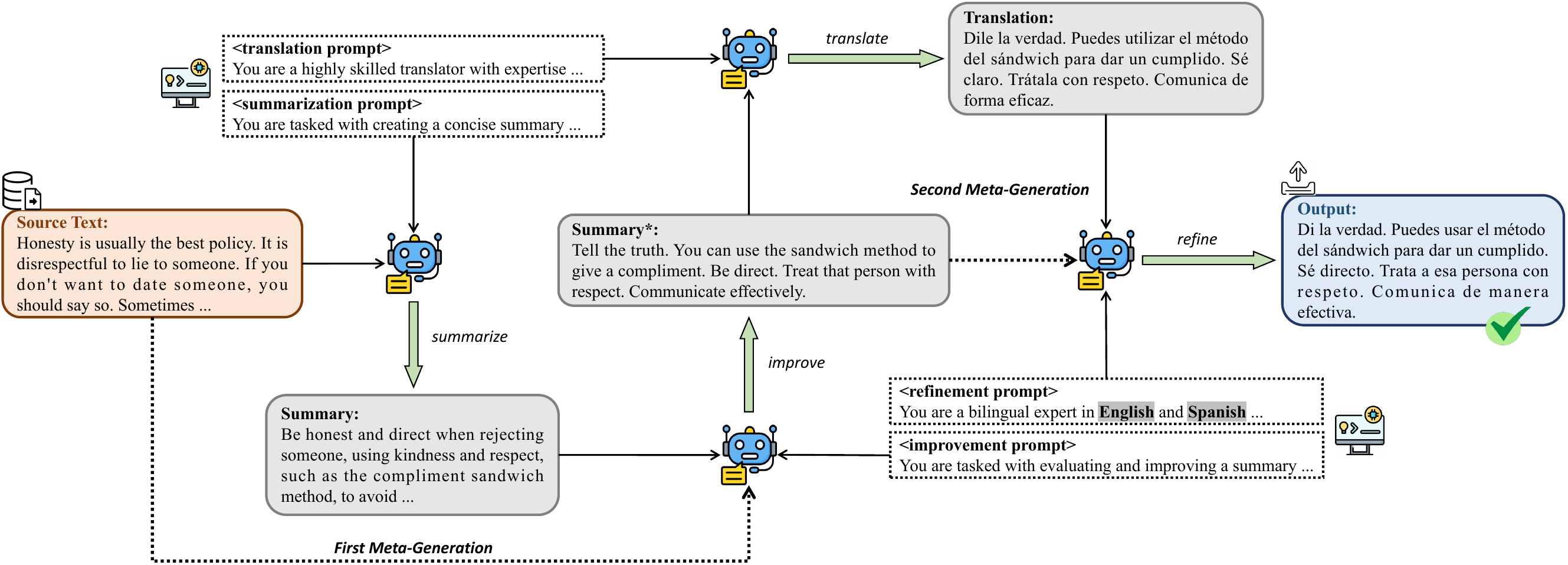}
	\caption{The architecture of our four-step zero-shot {\sitr} method for cross-lingual summarization.
		\label{fig:ourmethod}}
\end{figure*}

We test our method using \gptthreepointfive and \gptfour on two major cross-lingual summarization datasets, comparing them with fine-tuned models like mBART-50 \cite{mbart50tang}, mT5 \cite{mt5}, and other LLM baselines employing few-shot and summarize-translate approaches. Results show that \gptthreepointfive and \gptfour significantly outperform other LLM baselines across four metrics and even surpass fine-tuned models on most low-resource languages, demonstrating LLMs' strong capability in cross-lingual summarization under our proposed method.

We also apply our method to mainstream LLMs such as {\llamathree} \cite{llama3}, {\gemmatwo} \cite{gemma}, {\mixtral} \cite{mixtral87b}, and {\qwenonepointfive} \cite{qwen}, observing that most achieve impressive scores against the powerful {\gptfouro} model. This further validates the effectiveness of our approach and highlights that today's LLMs are capable of performing well on cross-lingual summarization tasks, even with low-resource languages.

In summary, this paper has the following contributions:
\begin{itemize}
\item Our proposed zero-shot {\sitr} method enhances large language models' performance on cross-lingual summarization for low-resource languages, demonstrating strong robustness across different models, datasets, and target languages.
\item Extensive experiments on two datasets and various low-resource languages reveal that our method significantly outperforms other LLM baselines and surpasses fine-tuned models.
\item To the best of our knowledge, we are the first to evaluate various LLMs on cross-lingual summarization for low-resource languages, showing that they possess the capability to achieve impressive results in this domain.
\end{itemize}

\section{Methodology}

\subsection{{\sitr} (Two-Stage Meta-Generation)}

In this paper, we propose a four-step zero-shot {\sitr} method for cross-lingual summarization in low-resource languages (see Figure \ref{fig:ourmethod}), comprising \textbf{{\summarization}}, \textbf{{\improvement}}, \textbf{{\translation}} and \textbf{{\refinement}}. The \textbf{{\improvement}} and \textbf{{\refinement}} stages align with two-stage meta-generation, involving LLM strategies like feedback learning, and rethinking \cite{metageneration}. To maximize LLMs' potential, we design specific prompts for each step, guiding the models to generate reliable outputs and minimizing error accumulation.

\stitle{[{\summarization}]} LLMs should distill the long input source text $ (\mathrm{I}) $ into concise summary $ (\mathrm{S}) $. To counter their tendency to generate overly detailed summaries, we use a summarization prompt $ (\mathrm{P_{\text{sum}}}) $ (see Figure \ref{fig: summarization_prompt}) to focus their output on the core essence of the text, ensuring the summary is both precise and relevant without unnecessary elaboration.
\begin{equation}
\mathrm{S} = \mathrm{LLM}(\mathrm{I};\,\mathrm{P_{\text{sum}}})
\end{equation}

\stitle{[{\improvement}]}
The first stage of meta-generation, providing large language models with the input source text $ (\mathrm{I}) $, the initial summary $ (\mathrm{S}) $ from the \textbf{{\summarization}} step, and the improvement prompt $ (\mathrm{P_{\text{imp}}}) $ (see Figure \ref{fig: improvement_prompt}) to recheck and optimize the summary $ (\mathrm{S^*}) $. This step reduces error accumulation by enabling self-improvement, preparing the more accurate summary for the next step of translation.
\begin{equation}
\mathrm{S^*} = \mathrm{LLM}(\mathrm{I};\,\mathrm{S};\,\mathrm{P_{\text{imp}}})
\end{equation}

\begin{table*}[!tb]
	\centering
	\begin{adjustbox}{width=0.75\linewidth}
    \renewcommand{\arraystretch}{1}
        \scriptsize
        \setlength{\tabcolsep}{4pt}
        \resizebox{\linewidth}{!}{
    		\begin{tabular}{@{}ccccc@{}}
    		\toprule
            \textbf{\,\,\,\,\,\,\,\,Dataset\,\,\,\,} & \textbf{Src Lang.} & \textbf{Trg Lang.} & \textbf{Domain} & \textbf{\,\,\,\,Train / Validation / Test\,\,\,\,} \\ \midrule \midrule
            \multirow{2}{*}{\quad CrossSum} & \multirow{2}{*}{English} & Uk \& Bn \& Id \& Gu & \multirow{2}{*}{News} & 1000 / 150 / 50 \\ \cmidrule(lr){3-3} \cmidrule(l){5-5} 
             &  & Pa &  & 769 / 100 / 50 \\ \midrule
            \quad WikiLingua & English & \,\,Id \& Vi \& Ar \& Hi \& Th\,\, & How-to Guide & 1000 / 150 / 50 \\ \bottomrule
    		\end{tabular}
            \vspace{-2mm}
        }
	\end{adjustbox}
	\caption{The source and amount of experimental data. The abbreviations of the languages correspond to their full names: Uk(Ukarainian), Bn(Bengali), Id(Indonesian), Gu(Gujarati), Pa(Pashto), Vi(Vietnamese), Ar(Arabic), Hi(Hindi), Th(Thai).\label{tab:data}}
\end{table*}

\stitle{[{\translation}]} 
Using the translation prompt $ (\mathrm{P_{\text{tra}}}) $ (see Figure \ref{fig: translation_prompt}), the optimized summary $ (\mathrm{S^*}) $ after the \textbf{{\improvement}} step is translated into the text $ (\mathrm{T}) $ in low-resource target language. Due to limited training data and lack of confidence, LLMs often produce redundant and messy outputs in these languages. This step aims to produce more reliable translations to ease the subsequent process.
\begin{equation}
\mathrm{T} = \mathrm{LLM}(\mathrm{S^*};\,\mathrm{P_{\text{tra}}})
\end{equation}

\stitle{[{\refinement}]} 
The optimized summary $ (\mathrm{S^*}) $ after the \textbf{{\improvement}} step, the initial translation $ (\mathrm{T}) $ from the \textbf{{\translation}} step, and the refinement prompt $ (\mathrm{P_{\text{ref}}}) $ (see Figure \ref{fig: refinement_prompt}) are combined and input into the LLMs for self-correction to generate the final output $ (\mathrm{O}) $. This process constitutes the second stage of meta-generation, enabling the LLMs to revise and produce a more accurate translation through re-evaluation.
\begin{equation}
\mathrm{O} = \mathrm{LLM}(\mathrm{S^*};\,\mathrm{T};\,\mathrm{P_{\text{ref}}})
\end{equation}

Our proposed method generally involves four steps to leverage the large language model's inherent capabilities. For summarization or translation tasks where a perfect result cannot be achieved in a single attempt, we utilize meta-generation to enable the LLMs to self-reflect and improve their final output. Additionally, when the model lacks guidance or confidence, we use strategic prompts to prevent disorganized or unreliable results. This approach ensures that the large language model produces high-quality and coherent outputs through the implementation of two-stage meta-generation.

\subsection{Large Language Models}
In this paper, we conduct a thorough evaluation of various large language models using our proposed \sitr method (Detail information in Appendix \ref{llmintroduction}).


\stitle{Closed-Source Models} We utilize four different models developed by OpenAI, including the latest {\gptfouro} and {\gptfouromini}.

\stitle{Open-Source Models} We conduct our experiments on {\llamathree} and {\llamathreepointone} \cite{llama2, llama3} developed by MetaAI; {\qwenonepointfive} and {\qwentwo} trained by Alibaba Cloud \cite{qwen}; {\gemma} and {\gemmatwo} created by Google \cite{gemma} and {\mixtral} from Mistral AI \cite{mixtral87b}.

\section{Experiments}

\subsection{Datasets \& Langueges}
\stitle{Datasets} In our research, we conduct experiments on two popular cross-lingual summarization datasets: \textbf{CrossSum} \cite{crosssum} and \textbf{Wikilingua} \cite{wikilingua}.



For fine-tuning experiments, we randomly select a subset from the training split of each dataset. For evaluation, we consistently use 50 randomly chosen samples from the test split to assess different methods and large language models.

\stitle{Languages} We consider the data ratio from the CommonCrawl corpus\footnote{\url{http://commoncrawl.org}} and its intersection with two datasets, aligning with the languages used in the previous study \cite{park}. Based on our research focus, we choose five challenging low-resource languages for each dataset to conduct our experiments. Detailed information about our experimental data is provided in Table \ref{tab:data} and more experimental languages are shown in Appendix \ref{morewxp}.

\begin{table*}[tb!]
	\centering
    \renewcommand{\arraystretch}{2.20}
    \fontsize{14pt}{14pt}\selectfont
        \setlength{\tabcolsep}{4pt}
        \resizebox{\linewidth}{!}{
    		\begin{tabular}{@{}ccccccccccccccccccccccccccc@{}}
    		\toprule
            \multicolumn{2}{c}{\multirow{3}{*}{\textbf{Model}}} & \multicolumn{20}{c}{\textbf{Language Pair}} & \multicolumn{5}{c}{\multirow{2}{*}{\textbf{Average Score}}} \\ \cmidrule{3-22}
            \multicolumn{2}{c}{} & \multicolumn{4}{c}{English\(\Rightarrow\)Ukrainian} & \multicolumn{4}{c}{English\(\Rightarrow\)Bengali} & \multicolumn{4}{c}{English\(\Rightarrow\)Indonesian} & \multicolumn{4}{c}{English\(\Rightarrow\)Gujarati} & \multicolumn{4}{c}{English\(\Rightarrow\)Pashto $^\spadesuit$} & \multicolumn{5}{c}{} \\ \cmidrule{3-27} 
            \multicolumn{2}{c}{} & R-1 & R-2 & R-L & BS & R-1 & R-2 & R-L & BS & R-1 & R-2 & R-L & BS & R-1 & R-2 & R-L & BS & R-1 & R-2 & R-L & BS & R-1 & R-2 & R-L & S-R & BS \\ \midrule \midrule
            \multirow{2}{*}{mBART-50} & 0-shot & 0.50 & 0.05 & 0.50 & 61.05 & 0.00 & 0.00 & 0.00 & 56.17 & 2.49 & 0.34 & 2.20 & 62.64 & 0.08 & 0.00 & 0.08 & 58.13 & 0.00 & 0.00 & 0.00 & 54.44 & 0.61 & 0.08 & 0.56 & 1.25 & 58.49 \\ 
             & 1000-shot & 13.46 & 3.17 & 10.03 & 67.64 & 9.74 & 1.74 & 8.86 & 67.60 & 18.20 & 3.23 & 14.36 & 68.87 & 11.23 & 1.56 & 10.26 & 67.97 & \textbf{23.63} & \textbf{5.36} & \textbf{16.41} & \textbf{74.12} & 15.25 & 3.01 & 11.98 & 30.24 & 69.24 \\ \midrule
            {mT5-small} & 1000-shot & 1.39 & 0.00 & 1.21 & 49.72 & 1.46 & 0.00 & 1.42 & 56.19 & 2.89 & 0.00 & 2.61 & 54.21 & 0.35 & 0.00 & 0.33 & 49.58 & 5.58 & 0.02 & 5.03 & 61.43 & 2.33 & 0.00 & 2.12 & 4.45 & 54.23 \\ \midrule
            {mT5-base} & 1000-shot & 1.72 & 0.00 & 1.57 & 49.39 & 1.76 & 0.07 & 1.67 & 60.43 & 3.72 & 0.00 & 3.47 & 56.73 & 1.09 & 0.00 & 1.09 & 50.74 & 6.11 & 0.05 & 5.50 & 61.66 & 2.88 & 0.02 & 2.66 & 5.56 & 55.79 \\ \midrule
            \multirow{3}{*}{{\gptthreepointfive} {\cite{park}}} & zero-shot & 10.83 & 1.26 & 6.78 & 65.74 & 5.69 & 0.60 & 2.95 & 60.93 & 11.36 & 1.93 & 7.58 & 66.19 & 6.90 & 1.21 & 3.97 & 65.05 & 3.25 & 0.37 & 2.77 & 60.87 & 7.61 & 1.07 & 4.81 & 13.49 & 63.76 \\ 
             & one-shot & 12.34 & 1.97 & 6.66 & 66.28 & 7.65 & 1.15 & 4.41 & 62.70 & 14.17 & 3.40 & 9.38 & 68.09 & 6.88 & 1.32 & 4.60 & 66.06 & 7.13 & 0.50 & 5.92 & 64.56 & 9.63 & 1.67 & 6.19 & 17.49 & 65.54 \\ 
             & two-shot & 13.48 & 1.51 & 6.57 & 66.02 & 8.50 & 0.89 & 5.47 & 65.65 & 14.38 & 3.12 & 10.19 & 68.21 & 9.47 & 1.64 & 6.68 & 67.10 & 6.23 & 0.66 & 5.41 & 63.84 & 10.41 & 1.56 & 6.86 & 18.83 & 66.16 \\ \midrule
            \multirow{3}{*}{{\gptfour} {\cite{park}}} & zero-shot & 8.75 & 1.91 & 5.75 & 65.35 & 8.51 & 1.31 & 5.74 & 65.21 & 8.94 & 1.84 & 6.14 & 65.70 & 8.14 & 1.09 & 6.00 & 66.85 & 10.10 & 2.06 & 7.29 & 68.27 & 8.89 & 1.64 & 6.18 & 16.71 & 66.28 \\ 
             & one-shot & 13.74 & 2.47 & 8.70 & 67.76 & 10.25 & 1.25 & 6.04 & 66.34 & 10.55 & 1.80 & 6.19 & 66.84 & 9.94 & 1.60 & 5.94 & 67.61 & 13.48 & 2.79 & 9.83 & 69.08 & 11.59 & 1.98 & 7.34 & 20.91 & 67.53 \\ 
             & two-shot & 13.40 & 2.25 & 8.42 & 67.87 & 10.97 & 1.96 & 7.23 & 67.09 & 12.60 & 2.34 & 8.57 & 67.41 & 10.20 & 1.80 & 6.23 & 68.67 & 14.63 & 2.92 & 9.30 & 68.82 & 12.36 & 2.25 & 7.95 & 22.56 & 67.97 \\ \midrule
            {{\gptthreepointfive} w/ summarize-translate} & zero-shot & 13.32 & 2.40 & 9.03 & 67.54 & 9.81 & 0.85 & 6.62 & 64.81 & 12.72 & 2.33 & 7.69 & 67.40 & 7.17 & 0.76 & 5.68 & 65.71 & 2.14 & 0.38 & 1.82 & 58.60 & 9.03 & 1.34 & 6.17 & 16.54 & 64.81 \\ 
            {{\gptfour} w/ summarize-translate} & zero-shot & 9.79 & 2.12 & 6.45 & 65.54 & 9.17 & 1.61 & 5.95 & 65.45 & 9.14 & 1.59 & 6.11 & 65.84 & 7.77 & 1.19 & 5.72 & 66.46 & 10.55 & 2.58 & 7.55 & 66.29 & 9.28 & 1.82 & 6.36 & 17.46 & 65.92 \\ \midrule
            {{\gptthreepointfive} w/ \sitr (Ours)} & zero-shot & \textbf{18.77} & \textbf{4.36} & \textbf{13.88} & \textbf{69.63} & \uline{14.28} & \uline{2.74} & \uline{10.16} & \textbf{69.47} & \uline{20.65} & \uline{4.44} & \textbf{15.57} & \textbf{69.16} & \textbf{14.58} & \textbf{2.58} & \textbf{12.06} & \textbf{70.41} & 17.06 & 2.56 & 13.84 & 71.29 & \uline{17.07} & \uline{3.34} & \textbf{13.10} & \uline{33.51} & \uline{69.99} \\ 
            {{\gptfour} w/ \sitr (Ours)} & zero-shot & \uline{17.04} & \uline{4.24} & \uline{11.74} & \uline{68.51} & \textbf{14.60} & \textbf{3.16} & \textbf{10.23} & \uline{69.39} & \textbf{20.86} & \textbf{4.69} & \uline{14.59} & \uline{69.05} & \uline{14.24} & \uline{2.45} & \uline{11.95} & \uline{70.28} & \uline{21.74} & \uline{5.23} & \uline{15.96} & \uline{73.43} & \textbf{17.70} & \textbf{3.95} & \uline{12.89} & \textbf{34.54} & \textbf{70.13} \\ \bottomrule
    		\end{tabular}
            \vspace{-2mm}
        }
	\caption{Experimental results on the CrossSum dataset. R-1, R-2, R-L, S-R and BS refer to ROUGE-1, ROUGE-2, ROUGE-L, sum of ROUGE-1/2/L and BERTScore respectively. The task with $^\spadesuit$ means training data less than 1000, where 1000-shot setting equals full fine-tuning, as the information shown in Table \ref{tab:data}. The best result on every target language is highlighted in \textbf{bold} font, and the second best result is marked with an \uline{underline}.}\label{tab:crosssummain}
\end{table*}

\begin{table*}[tb!]
	\centering
    \renewcommand{\arraystretch}{2.20}
    \fontsize{14pt}{14pt}\selectfont
        \setlength{\tabcolsep}{4pt}
        \resizebox{\linewidth}{!}{
		\begin{tabular}{@{}ccccccccccccccccccccccccccc@{}}
		\toprule
        \multicolumn{2}{c}{\multirow{3}{*}{\textbf{Model}}} & \multicolumn{20}{c}{\textbf{Language Pair}} & \multicolumn{5}{c}{\multirow{2}{*}{\textbf{Average Score}}} \\ \cmidrule{3-22}
        \multicolumn{2}{c}{} & \multicolumn{4}{c}{English\(\Rightarrow\)Indonesian} & \multicolumn{4}{c}{English\(\Rightarrow\)Vietnamese} & \multicolumn{4}{c}{English\(\Rightarrow\)Arabic} & \multicolumn{4}{c}{English\(\Rightarrow\)Hindi} & \multicolumn{4}{c}{English\(\Rightarrow\)Thai} & \multicolumn{5}{c}{} \\ \cmidrule{3-27} 
        \multicolumn{2}{c}{} & R-1 & R-2 & R-L & BS & R-1 & R-2 & R-L & BS & R-1 & R-2 & R-L & BS & R-1 & R-2 & R-L & BS & R-1 & R-2 & R-L & BS & R-1 & R-2 & R-L & S-R & BS \\ \midrule \midrule
        \multirow{2}{*}{mBART-50} & 0-shot & 2.48 & 0.22 & 1.93 & 63.74 & 0.71 & 0.08 & 0.66 & 63.43 & 0.23 & 0.05 & 0.21 & 61.77 & 1.46 & 0.34 & 1.37 & 58.95 & 11.14 & 1.14 & 10.09 & 59.27 & 3.20 & 0.37 & 2.85 & 6.42 & 61.43 \\ 
         & 1000-shot & 17.75 & 3.86 & 12.13 & 68.55 & 13.68 & 3.88 & 9.87 & 67.86 & 12.33 & 2.22 & 9.09 & 69.04 & 22.00 & 4.68 & 15.14 & 67.19 & 26.07 & 5.88 & 19.79 & 69.45 & 18.37 & 4.10 & 13.20 & 35.67 & 68.42 \\ \midrule
        {mT5-small} & 1000-shot & 0.52 & 0.00 & 0.50 & 51.87 & 0.36 & 0.00 & 0.36 & 55.24 & 0.00 & 0.00 & 0.00 & 55.77 & 8.54 & 0.25 & 7.95 & 63.61 & 5.58 & 0.07 & 4.44 & 53.81 & 3.00 & 0.06 & 2.65 & 5.71 & 56.06 \\ \midrule
        {mT5-base} & 1000-shot & 2.06 & 0.00 & 1.90 & 53.63 & 0.62 & 0.00 & 0.54 & 52.31 & 2.26 & 0.00 & 2.06 & 53.66 & 8.89 & 0.12 & 8.14 & 60.59 & 13.41 & 0.00 & 12.69 & 54.28 & 5.45 & 0.02 & 5.07 & 10.54 & 54.89 \\ \midrule
        \multirow{3}{*}{{\gptthreepointfive} {\cite{park}}} & zero-shot & 12.90 & 2.21 & 9.51 & 68.06 & 18.22 & 6.43 & 12.72 & 69.42 & 7.45 & 1.52 & 4.50 & 66.42 & 16.63 & 4.74 & 10.95 & 66.59 & 13.65 & 4.74 & 10.54 & 68.71 & 13.77 & 3.93 & 9.64 & 27.34 & 67.84 \\ 
         & one-shot & 16.26 & 3.47 & 10.90 & 69.08 & 22.77 & 8.61 & 16.62 & 71.14 & 9.94 & 1.84 & 6.77 & 68.20 & 17.53 & 4.35 & 11.03 & 67.45 & 14.18 & 4.61 & 10.79 & 68.74 & 16.14 & 4.58 & 11.22 & 31.94 & 68.92 \\ 
         & two-shot & 17.01 & 3.54 & 11.79 & 68.16 & 23.94 & 9.07 & 15.58 & 71.65 & 10.79 & 2.49 & 7.07 & 68.24 & 17.24 & \uline{5.11} & 12.72 & 68.56 & 18.12 & 4.73 & 11.40 & 69.77 & 17.42 & 4.99 & 11.71 & 34.12 & 69.28 \\ \midrule
        \multirow{3}{*}{{\gptfour} {\cite{park}}} & zero-shot & 13.75 & 2.98 & 9.78 & 67.81 & 16.44 & 6.40 & 11.83 & 68.34 & 8.49 & 1.45 & 5.21 & 66.46 & 16.76 & 4.32 & 10.60 & 66.62 & 19.28 & 5.86 & 14.86 & 69.47 & 14.94 & 4.20 & 10.46 & 29.60 & 67.74 \\ 
         & one-shot & 17.74 & 3.02 & 14.15 & 68.72 & 17.80 & 6.73 & 12.92 & 69.17 & 11.74 & 2.04 & 7.95 & 68.46 & 18.10 & 4.39 & 11.08 & 67.98 & 23.42 & 6.32 & 18.00 & 70.47 & 17.76 & 4.50 & 12.82 & 35.08 & 68.96 \\ 
         & two-shot & 18.03 & 3.05 & 13.26 & 68.89 & 20.31 & 6.44 & 13.22 & 70.48 & 13.21 & 2.80 & 9.22 & 69.93 & 19.79 & 5.01 & 12.88 & 68.28 & 24.35 & 6.21 & 18.91 & 70.21 & 19.14 & 4.70 & 13.50 & 37.34 & 69.56 \\ \midrule
         {{\gptthreepointfive} w/ summarize-translate} & zero-shot & 15.63 & 2.32 & 9.55 & 65.38 & 20.53 & 8.37 & 13.95 & 69.86 & 9.12 & 1.05 & 5.37 & 67.19 & 21.13 & 3.67 & 12.74 & 68.18 & 18.95 & 5.08 & 14.22 & 67.06 & 17.07 & 4.10 & 11.17 & 32.34 & 67.53 \\ 
        {{\gptfour} w/ summarize-translate} & zero-shot & 13.71 & 3.80 & 10.42 & 66.70 & 19.27 & 7.44 & 13.28 & 69.60 & 8.79 & 1.87 & 7.35 & 67.01 & 17.59 & 4.52 & 13.01 & 67.57 & 18.76 & 6.01 & 14.35 & 67.96 & 15.62 & 4.73 & 11.68 & 32.03 & 67.77 \\ \midrule
         {{\gptthreepointfive} w/ \sitr (Ours)} & zero-shot & \textbf{20.40} & \textbf{4.65} & \textbf{15.74} & \textbf{69.98} & \textbf{30.85} & \textbf{12.36} & \textbf{22.26} & \textbf{72.60} & \uline{14.38} & \uline{2.88} & \textbf{12.22} & \textbf{71.98} & \textbf{24.66} & \textbf{5.49} & \textbf{18.16} & \textbf{70.90} & \textbf{30.28} & \textbf{7.60} & \textbf{23.19} & \textbf{71.77} & \textbf{24.11} & \textbf{6.60} & \textbf{18.31} & \textbf{49.02} & \textbf{71.45} \\ 
        {{\gptfour} w/ \sitr (Ours)} & zero-shot & \uline{18.67} & \uline{3.94} & \uline{14.43} & \uline{69.12} & \uline{28.67} & \uline{10.57} & \uline{18.62} & \uline{72.08} & \textbf{15.32} & \textbf{3.67} & \uline{11.71} & \uline{71.06} & \uline{23.77} & 4.74 & \uline{16.86} & \uline{69.36} & \uline{28.76} & \uline{6.57} & \uline{20.59} & \uline{71.01} & \uline{23.04} & \uline{5.90} & \uline{16.44} & \uline{45.38} & \uline{70.53} \\ \bottomrule
		\end{tabular}
        \vspace{-2mm}
        }
	\caption{Experimental results on the WikiLingua dataset. R-1, R-2, R-L, S-R and BS refer to ROUGE-1, ROUGE-2, ROUGE-L, sum of ROUGE-1/2/L and BERTScore respectively. The best result on every target language is highlighted in \textbf{bold} font, and the second best result is marked with an \uline{underline}.}\label{tab:wikilinguamain}
\end{table*}

\subsection{Metrics}
In our experiments, we use \textbf{ROUGE-1/2/L} \cite{rouge} and \textbf{BERTScore} \cite{bertscore} as four different metrics. 

ROUGE metrics evaluate lexical overlap between the generated summaries and their references by considering unigrams, bigrams, and the longest common subsequence. 
BERTScore metric, however, focuses on measuring semantic similarity between two texts. We compute ROUGE scores with the multi-lingual ROUGE toolkit\footnote{\url{https://github.com/csebuetnlp/xl-sum/tree/master/multilingual_rouge_scoring}}, and BERTScore is calculated using the bert-score toolkit\footnote{\url{https://github.com/Tiiiger/bert_score}}.

\subsection{Baselines}
We select fine-tuned mBART-50, mT5-small, and mT5-base as baselines to demonstrate the capabilities of the fine-tuned encoder-decoder models on cross-lingual summarization tasks for low-resource languages.

For LLM-related baselines, we employ few-shot learning method with {\gptthreepointfive} and {\gptfour} following the prompt (see Figure \ref{fig: fewshot_prompt}) from a previous paper \cite{park}. Besides, we also evaluate the single-step summarize-translate method (see Figure \ref{fig: summarization_prompt}) as a baseline \cite{wangzeroshot}. (All implementation details are shown in Appendix \ref{expdetails}).




\subsection{Experiment Results}
The main experimental results on the CrossSum dataset are presented in Table \ref{tab:crosssummain}. We compare our zero-shot \sitr method with three types of baselines: fine-tuned encoder-decoder models, few-shot learning, and summarize-translate LLMs across various low-resource languages. Table \ref{tab:wikilinguamain} shows the main results for the WikiLingua dataset. (More experimental results are shown in Appendix \ref{morewxp}).

To further explore the potential of current large language models for cross-lingual summarization of low-resource languages and assess the robustness of our {\sitr} architecture, we conduct extensive experiments with our method on various large language models. The results are presented in Table \ref{tab:crosssumextra} and Table \ref{tab:wikilinguaextra}.

\stitle{{\sitr} vs Fine-tuned Models} 
Table \ref{tab:crosssummain} and Table \ref{tab:wikilinguamain} show that mT5-small and mT5-base both perform poorly on low-resource languages, even after fine-tuning with approximately 1,000 data points. While mBART-50 achieves better results, it still lags behind our zero-shot {\sitr} method across almost all languages, except for \textit{Pashto}, where fine-tuned mBART-50 has a slightly higher score. Notably, fine-tuning an encoder-decoder model for each low-resource language is significantly more costly than using large language models with our proposed \sitr method.

\begin{table*}[tb!]
	\centering
    \renewcommand{\arraystretch}{1.795}
    \fontsize{13pt}{13pt}\selectfont
        \setlength{\tabcolsep}{4pt}
        \resizebox{\linewidth}{!}{
		\begin{tabular}{@{}clcccccccccccccccccccccccc@{}}
	  \toprule
        \multicolumn{2}{c}{\multirow{3}{*}{\textbf{Model}}} & \multicolumn{20}{c}{\textbf{Language Pair}} & \multicolumn{4}{c}{\multirow{2}{*}{\textbf{Average Score}}} \\ \cmidrule{3-22}
        \multicolumn{2}{c}{} & \multicolumn{4}{c}{English\(\Rightarrow\)Ukrainian} & \multicolumn{4}{c}{English\(\Rightarrow\)Bengali} & \multicolumn{4}{c}{English\(\Rightarrow\)Indonesian} & \multicolumn{4}{c}{English\(\Rightarrow\)Gujarati} & \multicolumn{4}{c}{English\(\Rightarrow\)Pashto} & \multicolumn{4}{c}{} \\ \cmidrule{3-26} 
        \multicolumn{2}{c}{} & R-1 & R-2 & R-L & BS & R-1 & R-2 & R-L & BS & R-1 & R-2 & R-L & BS & R-1 & R-2 & R-L & BS & R-1 & R-2 & R-L & BS & R-1 & R-2 & R-L & BS \\ \midrule \midrule
        \multicolumn{2}{c}{\gptfouro} & 17.48 & 4.38 & 12.85 & 68.56 & \textbf{16.94} & \textbf{3.96} & \textbf{11.84} & \textbf{69.86} & \uline{22.42} & 5.03 & 16.37 & 69.64 & \textbf{15.76} & \textbf{3.59} & \textbf{13.04} & \textbf{71.09} & \textbf{22.80} & \uline{4.38} & \textbf{17.12} & \textbf{75.34} & \textbf{19.08} & \textbf{4.27} & \textbf{14.24} & \textbf{70.90} \\
        \multicolumn{2}{c}{\gptfouromini} & 17.41 & 4.38 & 12.20 & 68.73 & 15.34 & 2.90 & 10.26 & 69.01 & 18.54 & 3.04 & 13.59 & 69.78 & 13.49 & 2.60 & 11.16 & 70.58 & \uline{21.99} & \textbf{4.75} & \uline{15.68} & \uline{73.30} & 17.35 & 3.53 & 12.58 & 70.28 \\ \midrule
        \rowcolor{lightblue}
        \multicolumn{2}{c}{\llamaapimixtral} & 17.06 & \uline{4.63} & 12.52 & 68.48 & 12.48 & 1.92 & 9.65 & 68.61 & 14.28 & 3.47 & 10.35 & 68.64 & 8.35 & 0.76 & 6.82 & 68.60 & 6.51 & 0.63 & 5.11 & 63.36 & 11.74 & 2.28 & 8.89 & 67.54 \\
        \rowcolor{blue}
        \multicolumn{2}{c}{\groqmixtral} & 15.73 & 3.75 & 11.49 & 68.85 & 6.49 & 0.73 & 4.91 & 63.18 & 18.14 & 3.19 & 13.72 & 70.80 & 6.63 & 0.29 & 5.69 & 66.27 & 9.22 & 0.40 & 7.73 & 64.92 & 11.24 & 1.67 & 8.71 & 66.80 \\ \midrule
        \rowcolor{lightblue}
        \multicolumn{2}{c}{\llamaapiqwentst} & \textbf{18.02} & 4.28 & 12.02 & \uline{69.73} & 14.80 & 2.50 & 10.78 & 69.29 & 20.32 & 5.18 & 14.87 & 71.19 & 9.44 & 1.16 & 7.51 & 68.68 & 13.31 & 0.90 & 10.19 & 67.34 & 15.18 & 2.80 & 11.07 & 69.25 \\
        \rowcolor{lightblue}
        \multicolumn{2}{c}{\llamaapiqwenoooz} & 12.76 & 2.83 & 10.08 & 66.77 & 7.51 & 1.04 & 5.32 & 65.54 & 15.98 & 3.58 & 11.75 & 69.08 & 8.22 & 0.92 & 6.82 & 68.01 & 11.81 & 1.19 & 8.88 & 65.43 & 11.26 & 1.91 & 8.57 & 66.97 \\
        \rowcolor{lightblue}
        \multicolumn{2}{c}{\llamaapiqwenost} & 17.70 & 3.34 & 12.47 & 68.86 & 9.84 & 1.30 & 7.45 & 67.42 & 21.79 & 4.51 & 15.35 & 71.53 & 10.96 & 1.16 & 9.01 & 69.02 & 11.40 & 1.33 & 9.31 & 65.91 & 14.34 & 2.33 & 10.72 & 68.55 \\ \midrule
        \rowcolor{blue}
        \multicolumn{2}{c}{\groqllamate} & 9.92 & 2.29 & 7.08 & 65.40 & 9.83 & 1.78 & 6.67 & 65.86 & 14.59 & 3.28 & 11.32 & 68.08 & 10.21 & 1.77 & 8.76 & 67.23 & 6.52 & 0.90 & 5.57 & 62.67 & 10.21 & 2.00 & 7.88 & 65.85 \\
        \rowcolor{blue}
        \multicolumn{2}{c}{\groqllamatoe} & 15.35 & 3.39 & 11.22 & 68.89 & 14.42 & 2.33 & 9.64 & 68.42 & 21.64 & 4.56 & 14.79 & 71.44 & 9.57 & 0.84 & 7.15 & 66.42 & 9.35 & 1.44 & 7.17 & 65.05 & 14.07 & 2.51 & 9.99 & 68.04 \\
        \rowcolor{blue}
        \multicolumn{2}{c}{\groqllamatsz} & 9.16 & 1.87 & 7.12 & 64.20 & 6.59 & 1.71 & 4.34 & 65.53 & 16.46 & 4.37 & 12.10 & 69.09 & 8.34 & 1.59 & 6.62 & 65.88 & 13.68 & 2.78 & 10.54 & 69.12 & 10.85 & 2.46 & 8.14 & 66.76 \\
        \rowcolor{blue}
        \multicolumn{2}{c}{\groqllamatosz} & \uline{18.01} & \textbf{4.74} & \textbf{13.51} & 69.70 & \uline{16.04} & \uline{3.61} & \uline{11.16} & 69.03 & 21.59 & \uline{5.48} & 15.17 & 71.18 & 14.03 & 3.25 & 11.23 & 70.25 & 18.45 & 2.87 & 13.89 & 72.46 & \uline{17.62} & \uline{3.99} & 12.99 & \uline{70.52} \\ \midrule
        \rowcolor{blue}
        \multicolumn{2}{c}{\groqgemmas} & 12.93 & 1.86 & 10.48 & 68.48 & 12.67 & 1.44 & 9.45 & 68.95 & 20.23 & 3.44 & 15.78 & 71.29 & 7.67 & 0.89 & 6.60 & 67.54 & 1.46 & 0.00 & 1.46 & 58.71 & 10.99 & 1.53 & 8.75 & 66.99 \\
        \rowcolor{blue}
        \multicolumn{2}{c}{\groqgemmatn} & 16.93 & 3.84 & 13.24 & 69.37 & 13.47 & 1.82 & 9.57 & 69.27 & 21.62 & 4.89 & \uline{16.43} & \uline{71.98} & 14.11 & 3.10 & 11.60 & 70.39 & 13.13 & 1.19 & 9.98 & 68.47 & 15.85 & 2.97 & 12.16 & 69.90 \\
        \rowcolor{lightblue}
        \multicolumn{2}{c}{\llamaapigemmatts} & 17.08 & 4.03 & \uline{13.31} & \textbf{69.87} & 15.28 & 2.71 & 10.49 & \uline{69.81} & \textbf{23.28} & \textbf{5.58} & \textbf{17.84} & \textbf{72.50} & \uline{14.72} & \uline{3.36} & \uline{12.04} & \uline{70.97} & 16.18 & 1.66 & 11.38 & 71.35 & 17.31 & 3.47 & \uline{13.01} & \textbf{70.90} \\ \bottomrule
		\end{tabular}
        \vspace{-2mm}
        }
	\caption{Performance of various LLMs using our proposed {\sitr} method on CrossSum dataset. R-1, R-2, R-L and BS refer to ROUGE-1, ROUGE-2, ROUGE-L and BERTScore respectively. \colorbox{lightblue}{Light blue} and \colorbox{blue}{blue} denotes models inference through llama-api and groq. The best result on every target language is highlighted in \textbf{bold} font, and the second best result is marked with an \uline{underline}.}\label{tab:crosssumextra}
\end{table*}

\begin{table*}[tb!]
	\centering
    \renewcommand{\arraystretch}{1.795}
    \fontsize{13pt}{13pt}\selectfont
        \setlength{\tabcolsep}{4pt}
        \resizebox{\linewidth}{!}{
		\begin{tabular}{@{}clcccccccccccccccccccccccc@{}}
	  \toprule
        \multicolumn{2}{c}{\multirow{3}{*}{\textbf{Model}}} & \multicolumn{20}{c}{\textbf{Language Pair}} & \multicolumn{4}{c}{\multirow{2}{*}{\textbf{Average Score}}} \\ \cmidrule{3-22}
        \multicolumn{2}{c}{} & \multicolumn{4}{c}{English\(\Rightarrow\)Indonesian} & \multicolumn{4}{c}{English\(\Rightarrow\)Vietnamese} & \multicolumn{4}{c}{English\(\Rightarrow\)Arabic} & \multicolumn{4}{c}{English\(\Rightarrow\)Hindi} & \multicolumn{4}{c}{English\(\Rightarrow\)Thai} & \multicolumn{4}{c}{} \\ \cmidrule{3-26} 
        \multicolumn{2}{c}{} & R-1 & R-2 & R-L & BS & R-1 & R-2 & R-L & BS & R-1 & R-2 & R-L & BS & R-1 & R-2 & R-L & BS & R-1 & R-2 & R-L & BS & R-1 & R-2 & R-L & BS \\ \midrule \midrule
        \multicolumn{2}{c}{\gptfouro} & \textbf{22.31} & 4.63 & \textbf{16.25} & \textbf{70.31} & \textbf{30.95} & \uline{12.44} & \textbf{22.79} & \textbf{72.63} & \textbf{15.22} & \textbf{3.43} & \textbf{11.85} & \textbf{71.74} & \textbf{26.07} & 4.06 & 17.80 & \uline{69.78} & \textbf{30.87} & \textbf{9.06} & \textbf{23.38} & \textbf{72.52} & \textbf{25.08} & \uline{6.72} & \textbf{18.41} & \textbf{71.40} \\
        \multicolumn{2}{c}{\gptfouromini} & \uline{21.62} & 4.32 & \uline{16.05} & 69.94 & 28.12 & 11.56 & 19.20 & 71.74 & 13.44 & \uline{2.96} & 10.07 & 70.94 & 23.74 & 3.82 & 15.50 & 68.01 & \uline{29.11} & 6.04 & 21.05 & \uline{72.39} & 23.21 & 5.74 & 16.37 & \uline{70.60} \\ \midrule
        \rowcolor{lightblue}
        \multicolumn{2}{c}{\llamaapimixtral} & 19.54 & \textbf{5.70} & 15.20 & \uline{70.10} & 27.10 & 9.68 & 19.00 & 71.36 & 12.43 & 2.88 & \uline{10.36} & 70.23 & 21.28 & 4.57 & 15.44 & 67.86 & 27.44 & 7.31 & 20.83 & 70.29 & 21.56 & 6.03 & 16.17 & 69.97 \\
        \rowcolor{blue}
        \multicolumn{2}{c}{\groqmixtral} & 19.76 & 4.18 & 14.82 & 69.21 & 23.52 & 6.67 & 15.76 & 70.82 & 8.24 & 0.57 & 6.79 & 69.00 & 19.98 & 2.81 & 13.77 & 67.75 & 22.52 & 5.57 & 17.89 & 69.30 & 18.80 & 3.96 & 13.81 & 69.22 \\ \midrule
        \rowcolor{lightblue}
        \multicolumn{2}{c}{\llamaapiqwentst} & 17.54 & 3.84 & 13.21 & 68.68 & 30.09 & 11.40 & 21.08 & 72.22 & \uline{13.69} & 2.06 & 9.58 & \uline{71.21} & 22.96 & 3.52 & 15.08 & 67.74 & 30.25 & 8.49 & 22.40 & 71.42 & 22.91 & 5.86 & 16.27 & 70.25 \\
        \rowcolor{lightblue}
        \multicolumn{2}{c}{\llamaapiqwenoooz} & 19.24 & 5.19 & 14.47 & 69.37 & 29.89 & 10.81 & 20.19 & 72.37 & 12.38 & 2.09 & 9.51 & 70.49 & 25.57 & 4.74 & 17.62 & \textbf{70.12} & 26.43 & 6.75 & 19.77 & 70.17 & 22.70 & 5.92 & 16.31 & 70.50 \\
        \rowcolor{lightblue}
        \multicolumn{2}{c}{\llamaapiqwenost} & 19.76 & 3.81 & 14.37 & 69.67 & 30.66 & 12.23 & \uline{21.38} & \uline{72.60} & 12.41 & 2.50 & 10.16 & 70.20 & 21.89 & 2.26 & 14.71 & 68.69 & 28.82 & 7.46 & \uline{22.36} & 70.34 & 22.71 & 5.65 & 16.60 & 70.30 \\ \midrule
        \rowcolor{blue}
        \multicolumn{2}{c}{\groqllamate} & 16.78 & 4.02 & 12.75 & 67.63 & 22.91 & 7.79 & 15.89 & 69.77 & 9.39 & 1.21 & 7.82 & 67.62 & 20.86 & 4.59 & 15.43 & 67.26 & 23.21 & 5.85 & 18.28 & 68.02 & 18.63 & 4.69 & 14.03 & 68.06 \\
        \rowcolor{blue}
        \multicolumn{2}{c}{\groqllamatoe} & 20.34 & 4.67 & 14.15 & 69.63 & 29.92 & 10.71 & 20.49 & 72.20 & 10.20 & 0.89 & 8.69 & 68.79 & 24.26 & \uline{5.34} & 17.05 & 68.42 & 23.61 & 6.11 & 17.66 & 69.57 & 21.67 & 5.54 & 15.61 & 69.72 \\
        \rowcolor{blue}
        \multicolumn{2}{c}{\groqllamatsz} & 9.43 & 1.61 & 7.12 & 64.93 & 11.10 & 4.62 & 7.76 & 66.32 & 5.68 & 1.26 & 4.57 & 66.61 & 9.77 & 1.91 & 7.09 & 61.34 & 20.29 & 6.02 & 15.39 & 68.98 & 11.25 & 3.08 & 8.39 & 65.64 \\
        \rowcolor{blue}
        \multicolumn{2}{c}{\groqllamatosz} & 20.96 & \uline{5.04} & 14.27 & 69.79 & \uline{30.11} & \textbf{12.65} & 20.19 & 71.97 & 12.61 & 2.34 & 10.19 & 69.28 & \uline{26.04} & \textbf{6.52} & \textbf{19.33} & 69.72 & 29.01 & \uline{7.66} & 22.27 & 71.01 & \uline{23.75} & \textbf{6.84} & \uline{17.25} & 70.35 \\ \midrule
        \rowcolor{blue}
        \multicolumn{2}{c}{\groqgemmas} & 18.15 & 2.94 & 14.73 & 69.39 & 23.14 & 8.04 & 16.64 & 70.85 & 6.61 & 0.56 & 5.81 & 68.51 & 20.09 & 2.32 & 14.74 & 67.08 & 20.10 & 4.15 & 15.32 & 68.60 & 17.62 & 3.60 & 13.45 & 68.89 \\
        \rowcolor{blue}
        \multicolumn{2}{c}{\groqgemmatn} & 20.64 & 4.71 & 15.50 & 69.77 & 27.52 & 10.94 & 19.74 & 71.95 & 11.82 & 1.73 & 9.04 & 69.60 & 25.68 & 3.91 & 17.49 & 68.91 & 26.41 & 5.69 & 19.99 & 70.57 & 22.41 & 5.40 & 16.35 & 70.16 \\
        \rowcolor{lightblue}
        \multicolumn{2}{c}{\llamaapigemmatts} & 20.38 & 3.84 & 14.39 & 69.93 & 29.39 & 10.92 & 20.02 & 72.38 & 12.17 & 1.59 & 9.92 & 69.58 & 25.53 & 4.20 & \uline{18.03} & 68.91 & 27.95 & 5.66 & 20.93 & 70.68 & 23.08 & 5.24 & 16.66 & 70.30 \\ \bottomrule
		\end{tabular}
        \vspace{-2mm}
        }
	\caption{Performance of various LLMs using our proposed {\sitr} method on WikiLingua dataset. R-1, R-2, R-L and BS refer to ROUGE-1, ROUGE-2, ROUGE-L and BERTScore respectively. \colorbox{lightblue}{Light blue} and \colorbox{blue}{blue} denotes models inference through llama-api and groq. The best result on every target language is highlighted in \textbf{bold} font, and the second best result is marked with an \uline{underline}.}\label{tab:wikilinguaextra}
\end{table*}

\stitle{{\sitr} vs LLM Baselines} 
Table \ref{tab:crosssummain} and Table \ref{tab:wikilinguamain} demonstrate that under our approach, the outputs of the large language models significantly outperform other baselines in terms of both ROUGE and BERTScore metrics. This demonstrates that our outputs not only capture the key information of the text but also show notable improvements in word choice and semantic information.

On the CrossSum dataset, {\sitr} improves the sum of ROUGE-1/2/L scores from 18.83 to 33.51 (a 78\% increase) with {\gptthreepointfive} and from 22.56 to 34.54 (a 53\% increase) with {\gptfour}, compared to two-shot generation. The improvement over the summarize-translate method is even more notable, with the sum of ROUGE-1/2/L scores increasing by 103\% (from 16.54 to 33.51) with {\gptthreepointfive} and 98\% (from 17.46 to 34.54) with {\gptfour}.

\begin{figure*}[!tb]
	\centering
	\includegraphics[width=1.00\linewidth]{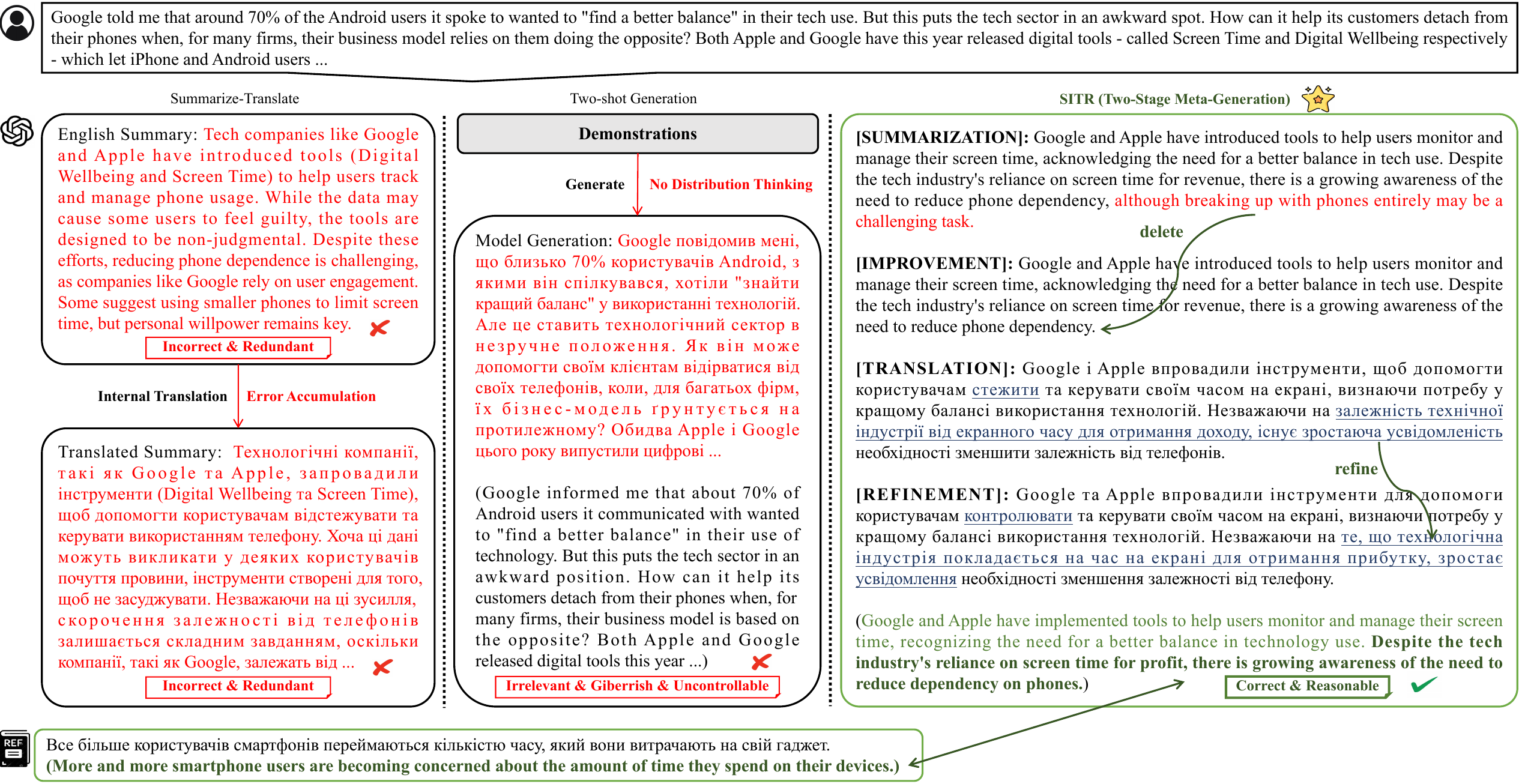}
	\caption{Comparison of three different LLM methods on one single test example to summarize English source text in Ukarainian. The English translation of each model output is shown in brackets. 
    \vspace{-2mm}
\label{fig:comparison}}
\end{figure*}

On the WikiLingua dataset, the sum of ROUGE-1/2/L scores improves by 44\% (from 34.12 to 49.02) and 21\% (from 37.34 to 45.38) when comparing two-shot generation to our zero-shot method. Additionally, the increases are 52\% (from 32.34 to 49.02) and 42\% (from 32.03 to 45.38) when compared to the summarize-translate method.

For BERTScore, our method shows an increase of 2 to 3 percentage points compared to two-shot generation, and 4 to 5 percentage points compared to the summarize-translate method. This significant improvement reflects a substantial enhancement in the semantic quality of the model's outputs.

\stitle{Improvement via Language} 
According to Table \ref{tab:crosssummain} and Table \ref{tab:wikilinguamain}, our proposed {\sitr} method achieves a smaller percentage improvement on the WikiLingua dataset compared to CrossSum. This discrepancy may stem from the fact that while the target languages in WikiLingua are still low-resource, they are relatively more studied, providing greater resources for pre-trained LLMs. As a result, our method tends to yield more significant improvements in languages with fewer available resources. (Additional results can be found in Appendix \ref{morewxp}.)

\stitle{Robustness of \sitr and Capabilities of LLMs} 
Table \ref{tab:crosssummain} and Table \ref{tab:wikilinguamain} demonstrate the impressive performance of \gptthreepointfive and \gptfour using our \sitr method.
Additionally, Table \ref{tab:crosssumextra} and Table \ref{tab:wikilinguaextra} show how our \sitr method effectively leverages large language models, allowing many of them to excel in cross-lingual summarization, even for low-resource languages. Notable examples include high-performing open-source models like \llamathreeoneseventyb and \gemmatwosevenb.

When comparing Table \ref{tab:crosssummain} and Table \ref{tab:crosssumextra}, it becomes evident that many open-source large language models, under our \sitr method, significantly outperform \gptfour using two-shot learning. Table \ref{tab:wikilinguaextra} further reveals that while \gptfouro consistently leads in most metrics, the other models achieve second-best performances across various languages and metrics. Overall, many LLMs could deliver impressive results on average, which also demonstrates the robustness of our \sitr method.

\stitle{Parameter via Capability} 
From Table \ref{tab:crosssumextra} and Table \ref{tab:wikilinguaextra}, we observe that a large language model's cross-lingual capabilities on low-resource languages are not solely dependent on the number of model parameters. For example, within the {\qwenonepointfive} series, {\qwenseventwob} outperforms {\qwenoneonezerob} in several low-resource languages, such as \textit{Gujarati} and \textit{Ukrainian}. Additionally, the {\gemmatwonineb} and {\gemmatwosevenb} models demonstrate strong performance, with {\gemmatwosevenb} achieving the best result in \textit{Indonesian} and the second-best result in \textit{Gujarati}, only slightly behind the {\gptfouromini} model on average.

\stitle{Best LLM Under Our Method}
We evaluate the performance of various LLMs, with the results shown in Table \ref{tab:crosssumextra} and Table \ref{tab:wikilinguaextra}. Overall, the GPT series models consistently deliver better performance across different low-resource languages, particularly {\gptfouro}, which is considered one of the most powerful LLMs available. While other open-source models do not surpass the overall performance of {\gptfouro}, some are able to match or even exceed its performance in specific languages.

\subsection{Output Analysis}

In Figure \ref{fig:comparison}, we compare our {\sitr} method with other two LLM baselines in summarizing English news into Ukrainian. 

The outputs from the other two methods are suboptimal due to their lack of relevance to the main topic and the generation of nonsensical content. The single-step summarize-translate method, which lacks self-correction and crucial prompt guidance, translates inaccurate summaries directly into the target language, causing error accumulation. On the other hand, the two-shot generation method skips the distributed thinking process, leading to uncontrollable outputs when the model fails to learn effectively from the examples. Both approaches, therefore, exhibit significant limitations.

In contrast, our method leverages meta-generation with targeted guidance, ensuring the model produces controlled and coherent outputs. This approach also allows the model to engage in self-reflection and iterative improvement, leading to more reliable and accurate results. 
The improvement step streamlines the summary by removing unnecessary sentences, while the refinement step adjusts sentence structure to better match the style of news reporting. 
Compared to the dataset's reference, our method captures the essence of the source text even more effectively.

\subsection{Ablation Studies}
Our proposed method improves upon the traditional single-step summarize-translate approach by integrating tailored prompts and employing a two-stage meta-generation process, which involves enhancing the summary and refining the translation.

The two additional steps, \textbf{{\improvement}} and \textbf{{\refinement}}, utilize meta-generation to optimize output and minimize error accumulation. These distinctions are particularly critical for cross-lingual summarization tasks in low-resource languages. Thus, we pose the question: \textit{How significantly do meta-generation steps impact the overall performance of LLMs on this task?}

\begin{figure}[!tb]
	\centering
	\includegraphics[width=1\linewidth]{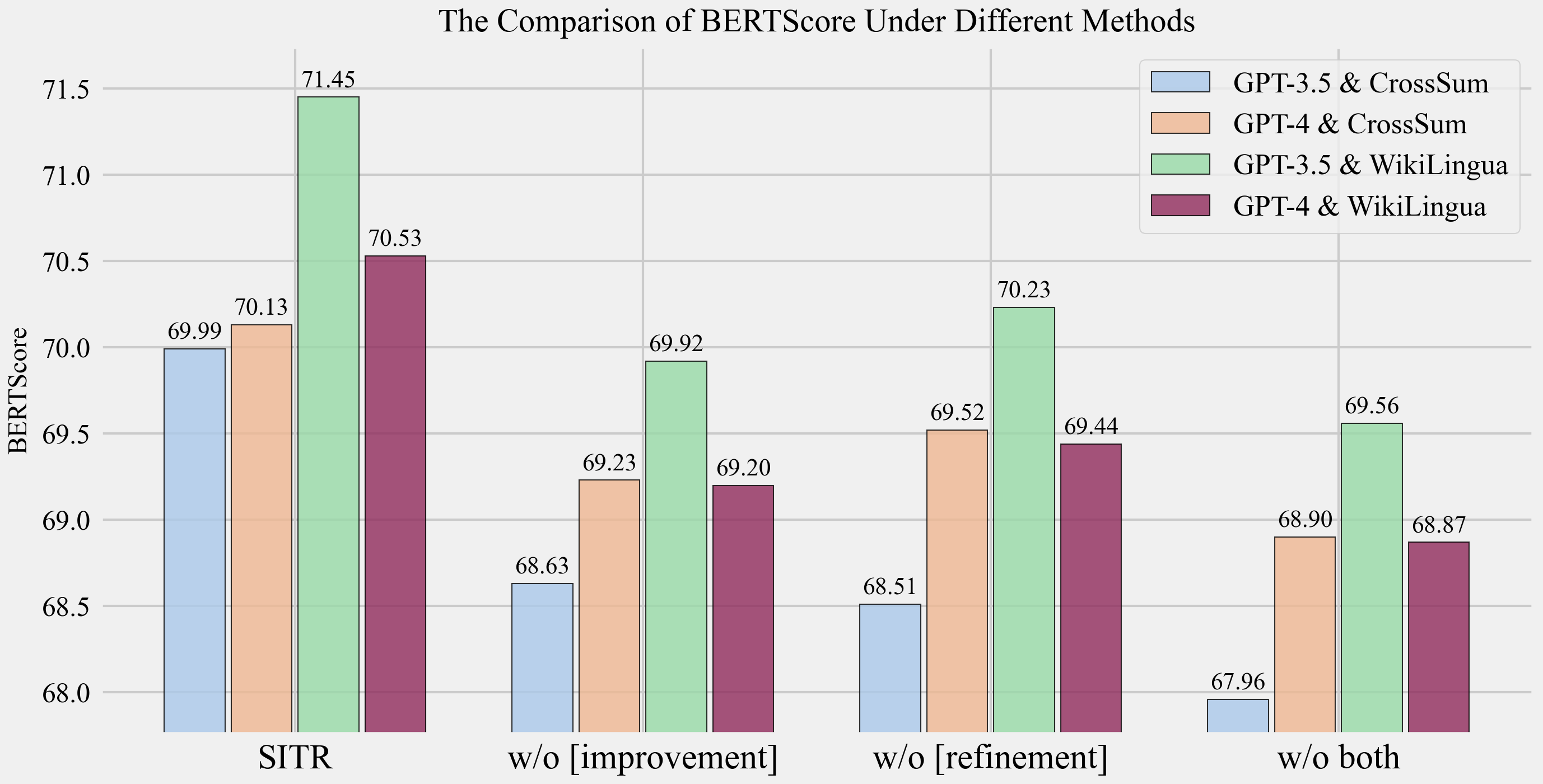}
	\caption{Comparison of the BERTScore after removing key meta-generation steps.
    \vspace{-2mm}
\label{fig:ablationbertscore}}
\end{figure}

\begin{figure}[!tb]
	\centering
	\includegraphics[width=1\linewidth]{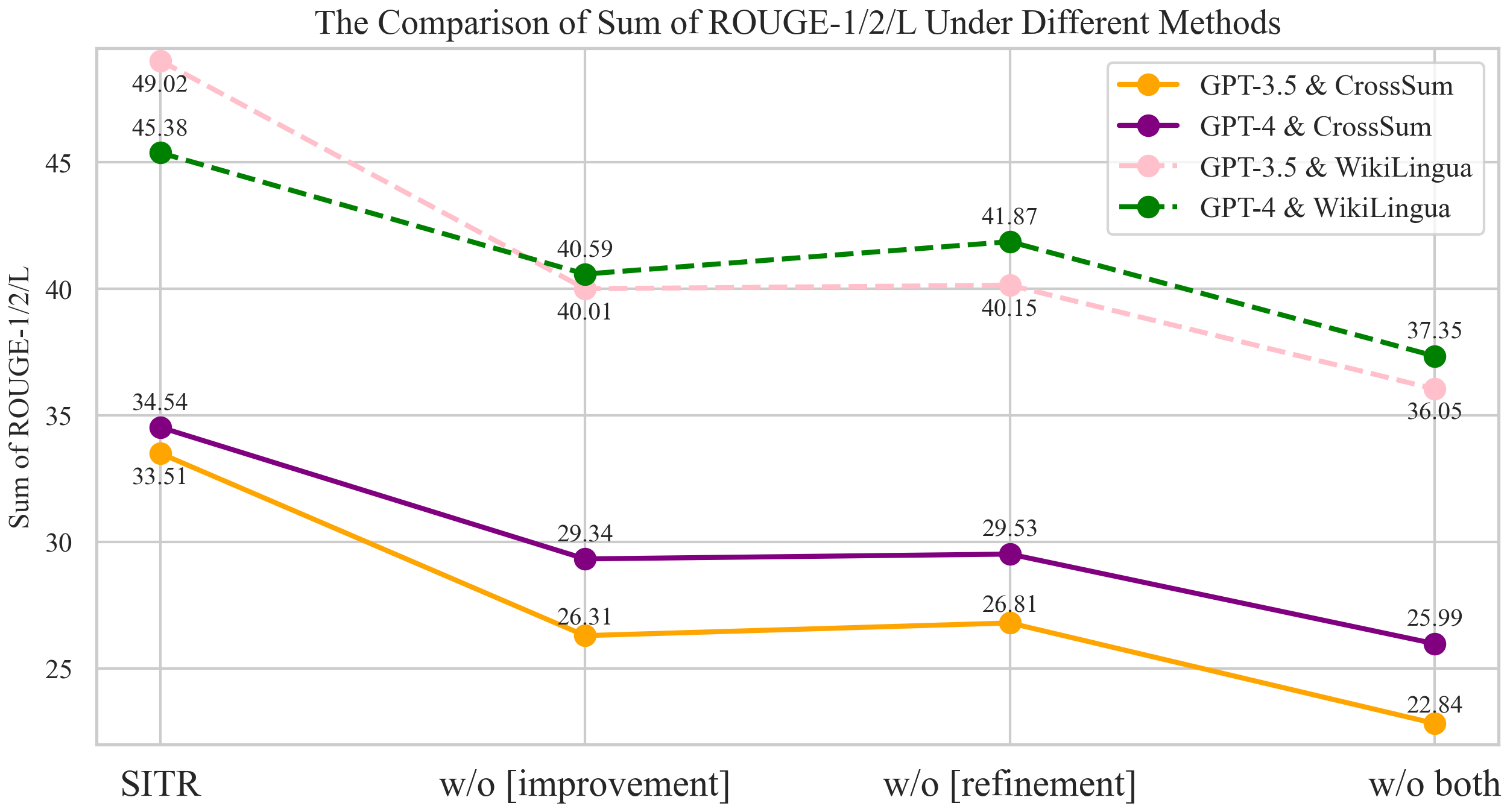}
	\caption{Comparison of the sum of ROUGE-1/2/L after removing key meta-generation steps.
    \vspace{-2mm}
\label{fig:ablationsumrouge}}
\end{figure}

Here, we carry out three sets of comparative experiments to demonstrate the importance of two meta-generation steps: (i) Delete the \textbf{{\improvement}} step. (ii) Delete the \textbf{{\refinement}} step. (iii) Delete both the \textbf{{\improvement}} and \textbf{{\refinement}} step.

The ablation experimental results across different metrics are presented in Figure \ref{fig:ablationbertscore} and Figure \ref{fig:ablationsumrouge}, where we compare the performance of the complete {\sitr} architecture with the three ablation experiments. 

From these results, we could find that each step positively impacts the final outcomes. Deleting any step results in decreased ROUGE-1/2/L and BERTScore metrics, underscoring the overall significance of our proposed {\sitr} architecture. Specifically, deleting the \textbf{{\improvement}} step results in an approximate 18.4\% drop in the sum of ROUGE-1/2/L scores (from 49.02 to 40.01) for {\gptthreepointfive}. Also, the BERTScore would decrease by 1-2 percentage points, which indicates a noticeable loss in semantic quality.

In summary, our proposed {\sitr} method illustrates the cooperative and complementary nature of its architecture.
This demonstrates the robustness of our {\sitr} method: when model ouputs are less than ideal, the \textbf{{\improvement}} and \textbf{{\refinement}} steps allow the model to self-correct and reassess, mitigating the impact of error accumulation on the final output. (Further studies in Appendix \ref{moreablation}).

\section{Related Works}
Cross-lingual summarization is a critical task in natural language processing, involving the generation of a summary for text in one language based on a source text in another language \cite{cls_survey, longdoccls}. The emergence of deep learning-based neural machine translation systems \cite{nmt1, nmt2, nmt3, nmt4} and text summarization approaches \cite{rnntextsum1, rnntextsum2, rnntextsum3}, particularly those leveraging recurrent neural networks \cite{rnn, gru, lstm}, enhanced model performance on CLS tasks. 

Later, advances in neural network technologies, especially the Transformer architecture \cite{transformer}, have led to the development of end-to-end CLS models that integrate translation and summarization into a single framework, improving overall performance.
Recent years, large language models have experienced a period of rapid development and widespread adoption \cite{instructgpt, gpt3, llama2}, and they have gained attention for their potential in cross-lingual summarization. \citet{wangzeroshot} showed their strong capabilities in high-resource languages like Chinese and German, while \citet{park} found that LLMs using few-shot approaches still struggle with low-resource languages. This investigation is crucial for understanding and improving the models' ability to produce accurate and coherent summaries across various languages, thereby expanding the scope and applicability of LLMs in the CLS domain.

\section{Conclusion and Future Work}
In this paper, we introduce a four-step zero-shot {\sitr} architecture, demonstrating the potential of LLMs for cross-lingual summarization in low-resource languages. Our approach enables LLMs to outperform three baseline types across various metrics, achieving notable performance in this domain.

We apply our {\sitr} method to evaluate a wide range of LLMs, revealing their strong performance in cross-lingual summarization for low-resource languages and further demonstrating the robustness of our approach. For future research, we plan to investigate more effective methodologies to further unlock the potential of LLMs in this domain. 

\section*{Limitations}
While we evaluate the performance of LLMs in cross-lingual summarization on two datasets to showcase the effectiveness of both our zero-shot \sitr method and the models, this study has several limitations:
(i) The design of prompts can affect model performance, partly due to the models' limited confidence with low-resource languages. Future research could explore methods to enable large language models to generate reliable outputs without depending on manually designed prompts.
(ii) We do not examine cross-lingual summarization tasks involving two low-resource languages. Future work could address this gap to fully explore the potential of LLMs in these more challenging scenarios.


\bibliography{custom}
\bibliographystyle{acl_natbib}

\newpage
\appendix

\begin{table*}[tb!]
	\centering
    \renewcommand{\arraystretch}{2}
    \fontsize{13pt}{13pt}\selectfont
        \setlength{\tabcolsep}{4pt}
        \resizebox{\linewidth}{!}{
		\begin{tabular}{@{}ccccccccccccccccccccccc@{}}
	  \toprule
        \multicolumn{2}{c}{\multirow{3}{*}{\textbf{Model}}} & \multicolumn{16}{c}{\textbf{Language Pair}} & \multicolumn{5}{c}{\multirow{2}{*}{\textbf{Average Score}}} \\ \cmidrule{3-18}
        \multicolumn{2}{c}{} & \multicolumn{4}{c}{English\(\Rightarrow\)Igbo $^\spadesuit$} & \multicolumn{4}{c}{English\(\Rightarrow\)Hausa $^\spadesuit$} & \multicolumn{4}{c}{English\(\Rightarrow\)Nepali} & \multicolumn{4}{c}{English\(\Rightarrow\)Yoruba $^\spadesuit$} & \multicolumn{5}{c}{} \\ \cmidrule{3-23} 
        \multicolumn{2}{c}{} & R-1 & R-2 & R-L & BS & R-1 & R-2 & R-L & BS & R-1 & R-2 & R-L & BS & R-1 & R-2 & R-L & BS & R-1 & R-2 & R-L & S-R & BS \\ \midrule \midrule
        \multirow{2}{*}{mBART-50} & 0-shot & 3.38 & 0.00 & 2.46 & 59.65 & 3.20 & 0.31 & 2.93 & 55.76 & 0.00 & 0.00 & 0.00 & 60.07 & 2.86 & 0.05 & 1.82 & 57.34 & 2.36 & 0.09 & 1.80 & 4.25 & 58.21 \\ 
         & 1000-shot & 6.59 & 0.00 & 5.50 & 60.89 & 14.24 & 2.94 & 11.57 & 67.34 & 9.40 & 1.91 & 8.18 & 68.18 & 7.68 & 0.00 & 7.43 & 61.10 & 9.48 & 1.21 & 8.17 & 18.86 & 64.38 \\ \midrule
         {mT5-base} & 1000-shot & 0.43 & 0.00 & 0.43 & 51.76 & 6.90 & 0.17 & 5.61 & 55.82 & 0.72 & 0.00 & 0.72 & 56.35 & 0.55 & 0.00 & 0.55 & 51.28 & 2.15 & 0.04 & 1.83 & 4.02 & 53.80 \\ \midrule
        \multirow{3}{*}{\gptthreepointfive \cite{park}} & zero-shot & 3.44 & 0.35 & 2.61 & 58.23 & 4.95 & 0.67 & 4.12 & 57.71 & 4.30 & 0.56 & 3.27 & 63.87 & 2.34 & 0.32 & 1.95 & 58.48 & 3.76 & 0.48 & 2.99 & 7.23 & 59.57 \\ 
         & one-shot & 4.35 & 0.59 & 3.27 & 58.45 & 7.82 & 0.88 & 6.21 & 61.78 & 5.26 & 0.40 & 4.74 & 66.24 & 2.87 & 0.15 & 2.01 & 58.72 & 5.08 & 0.51 & 4.06 & 9.65 & 61.30 \\ 
         & two-shot & 6.85 & 1.46 & 5.22 & 60.13 & 10.12 & 2.05 & 7.16 & 62.59 & 5.56 & 0.70 & 4.46 & 65.52 & 4.07 & 1.10 & 2.81 & 58.27 & 6.65 & 1.33 & 4.91 & 12.89 & 61.63 \\ \midrule
        \multirow{3}{*}{\gptfour \cite{park}} & zero-shot & 7.64 & 1.59 & 5.70 & 65.34 & 11.67 & 3.50 & 7.65 & 65.10 & 5.73 & 0.76 & 4.24 & 65.23 & 6.33 & 0.96 & 4.86 & 65.77 & 7.84 & 1.70 & 5.61 & 15.15 & 65.36 \\ 
         & one-shot & 9.24 & 2.09 & 6.66 & 65.90 & 12.97 & \uline{3.71} & 8.61 & 65.33 & 7.32 & 1.15 & 5.30 & 66.09 & 6.59 & 1.38 & 5.29 & 65.80 & 9.03 & 2.08 & 6.47 & 17.58 & 65.78 \\ 
         & two-shot & 9.58 & 1.99 & 6.94 & 65.64 & 12.84 & 3.54 & 8.70 & 65.49 & 7.21 & 1.66 & 5.57 & 66.35 & 6.76 & 1.00 & 5.21 & 65.76 & 9.10 & 2.05 & 6.61 & 17.76 & 65.81 \\ \midrule
         {\gptthreepointfive w/ summarize-translate} & zero-shot & 3.67 & 0.21 & 2.84 & 57.69 & 8.36 & 1.29 & 6.07 & 61.91 & 5.82 & 0.71 & 4.64 & 66.25 & 4.64 & 0.90 & 3.83 & 59.84 & 5.62 & 0.78 & 4.35 & 10.75 & 61.42 \\ 
        {\gptfour w/ summarize-translate} & zero-shot & 6.61 & 1.22 & 5.30 & 65.44 & 12.44 & 3.08 & 8.52 & 65.48 & 5.29 & 0.91 & 3.95 & 65.51 & 4.93 & 0.72 & 3.92 & 65.13 & 7.32 & 1.48 & 5.42 & 14.22 & 65.39 \\ \midrule
         {\gptthreepointfive w/ \sitr (Ours)} & zero-shot & \uline{15.56} & \uline{2.79} & \uline{10.63} & \uline{67.31} & \uline{19.60} & 2.15 & \uline{14.84} & \uline{69.33} & \uline{12.89} & \uline{2.51} & \uline{9.63} & \uline{69.69} & \uline{11.82} & \uline{2.20} & \uline{10.69} & \uline{68.41} & \uline{14.97} & \uline{2.41} & \uline{11.45} & \uline{28.33} & \uline{68.69} \\ 
        {\gptfour w/ \sitr (Ours)} & zero-shot & \textbf{20.47} & \textbf{4.02} & \textbf{14.90} & \textbf{70.17} & \textbf{23.02} & \textbf{4.30} & \textbf{14.22} & \textbf{70.04} & \textbf{15.81} & \textbf{3.17} & \textbf{11.09} & \textbf{70.19} & \textbf{15.39} & \textbf{2.15} & \textbf{9.81} & \textbf{70.90} & \textbf{18.67} & \textbf{3.41} & \textbf{12.51} & \textbf{34.59} & \textbf{70.33} \\ \bottomrule
		\end{tabular}
        \vspace{-2mm}
        }
	\caption{Additional experimental results on the CrossSum dataset. R-1, R-2, R-L, S-R and BS refer to ROUGE-1, ROUGE-2, ROUGE-L, sum of ROUGE-1/2/L and BERTScore respectively. The task with $^\spadesuit$ means training data less than 1000, where 1000-shot setting equals full fine-tuning. The best result on every target language is highlighted in \textbf{bold} font, and the second best result is marked with an \uline{underline}.}\label{tab:crosssumappendixone}
\end{table*}

\section{Large Language Models}\label{llmintroduction}
Here, we give the introduction of LLMs used in this paper.
\begin{itemize}
\item {\gptthreepointfive}: A powerful language model developed by OpenAI, designed to generate human-like text based on input prompts, which is highly effective for a variety of natural language processing (NLP) tasks. 
\item {\gptfour}: An advanced multi-modal LLM from OpenAI, which is able to accept both images and texts to do text-generation, and exhibits human-level performance on various NLP benchmarks.
\item {\gptfouro}: A better multi-modal LLM compared with {\gptfour} which was released on May 13, 2024 by OpenAI.
\item {\gptfouromini}: A most cost-efficient multi-modal small model released on July 18, 2024 by OpenAI, which enables a broad range of tasks with low cost and latency. 
\item {\llamathree}: A family of large language models includes two versions with 8B and 70B parameters developed by MetaAI, which were trained on 15 trillion tokens data.
\item {\llamathreepointone}: The latest and strongest open-source LLM family released by MetaAI, contains three models with 8B, 70B and 405B parameters.
\item {\qwenonepointfive}: The improved version of {\qwen}, the LLM family developed by Alibaba Cloud. The whole family contains eight models: 0.5B, 1.8B, 4B, 7B, 14B, 32B, 72B, and 110B.
\item {\qwentwo}: Newest model series from the Qwen team with better performances.
\item {\gemma}: Light-weight, text-to-text, decoder-only large language models trained by Google, which have two versions with 2B and 7B parameters.
\item {\gemmatwo}: The next generation of open-source models and improved version of {\gemma}, released by Google on June 27, 2024, which contains three versions with 2B, 9B and 27B parameters.
\item {\mixtral}: Mixture of Experts (MoE) models \cite{moe} with 8 experts trained by Mistral AI, now have two versions 8x7B and 8x22B.
\end{itemize}

\section{Experiments Details}\label{expdetails}
We primarily use {\gptthreepointfive}, {\gptfour}, {\gptfouro}, and {\gptfouromini} models via OpenAI's official API\footnote{\url{https://openai.com/}}. Additionally, we utilize llama-api\footnote{\url{https://www.llama-api.com/}} to access two models from the {\qwenonepointfive} family, {\qwentst}, {\mixtralone}, and {\gemmatts}. For the remaining models, we conduct experiments using the groq platform\footnote{\url{https://groq.com/}}.

For all LLM-related experiments, we set the \textit{temperature} to 0.0 and \textit{top-p} to 0.95 to minimize randomness and ensure consistent model outputs.
To reproduce the few-shot results from the previous paper with {\gptthreepointfive} and {\gptfour}, we use the provided prompt \cite{park} and access the OpenAI official API.

For baselines requiring further fine-tuning, we use three encoder-decoder transformer models: mBART-50\footnote{\url{https://huggingface.co/facebook/mbart-large-50}}, mT5-base\footnote{\url{https://huggingface.co/google/mt5-base}}, and mT5-small\footnote{\url{https://huggingface.co/google/mt5-small}}. If a low-resource language's training data exceeds 1,000 samples, we randomly select 1,000 for the 1,000-shot experiments. If the data contains fewer than 1,000 samples, we use all available data for fine-tuning. To fine-tune mBART-50, mT5-base, and mT5-small, we perform all experiments on a single 24GB-VRAM A5000 GPU. We set the training epochs to 3, with learning rates of 1e-4, 3e-4, and 5e-4, respectively, and select the checkpoint with the highest sum of ROUGE-1/2/L scores.

At last, for OpenAI's {\gptthreepointfive} and {\gptfour} models, we use the \textit{gpt-3.5-turbo-0125} and \textit{gpt-4-0125-preview} versions, conducting all experiments between July 15th and August 1st.

\section{Additional Experiments}\label{morewxp}

To better illustrate the effectiveness of our proposed \sitr method in leveraging large language models for cross-lingual summarization in low-resource languages, we select eight additional low-resource languages from the CrossSum dataset (Igbo, Hausa, Nepali, Yoruba, Oromo, Welsh, Urdu, Swahili) and sample 30 test examples for each language. We evaluate our method against fine-tuned models and other LLM baselines, with the results presented in Table \ref{tab:crosssumappendixone} and Table \ref{tab:crosssumappendixtwo}.

We observe that \gptthreepointfive and \gptfour still significantly and consistently outperform all other baselines when using our proposed \sitr method. 

The results from Table \ref{tab:crosssumappendixone} and Table \ref{tab:crosssumappendixtwo} show that, despite fine-tuning, the performance of the mT5-base model remains unsatisfactory, while the mBART-50 model performs significantly better. However, when comparing the scores of fine-tuned mBART-50 with our proposed zero-shot \sitr method, it is clear that \sitr still holds a significant advantage, showing notable improvements in both BERTScore and all three ROUGE metrics.

In Table \ref{tab:crosssumappendixone}, the sum of ROUGE-1/2/L scores for \gptthreepointfive improves by 124\% (from 12.89 to 28.83) and 168\% (from 10.75 to 28.83) when comparing our \sitr method with two-shot generation and the summarize-translate method. For \gptfour, the improvements are 95\% (from 17.76 to 34.59) and 143\% (from 14.22 to 34.59) respectively.

In Table \ref{tab:crosssumappendixtwo}, the sum of ROUGE-1/2/L scores for \gptthreepointfive increases by 58\% (from 25.59 to 40.35) and 84\% (from 21.90 to 40.35) when comparing our \sitr method with the best few-shot generation method (one-shot generation worked best for \gptthreepointfive) and the summarize-translate method. For \gptfour, the improvements are 51\% (from 26.89 to 40.64) and 112\% (from 19.20 to 40.64) respectively.

Additionally, our method shows significant improvement in the BERTScore metric, indicating a substantial semantic advantage for \sitr outputs. Moreover, we find that the improvement of few-shot learning methods is significantly constrained when transitioning from one-shot to two-shot. In fact, Table \ref{tab:crosssumappendixtwo} shows a slight performance decline with \gptthreepointfive, suggesting that few-shot learning may face certain limitations on this task.

These results, based on experiments with eight low-resource languages, further demonstrate that our \sitr method effectively harnesses the capabilities of large language models in this domain.

\begin{table*}[tb!]
	\centering
    \renewcommand{\arraystretch}{2}
    \fontsize{13pt}{13pt}\selectfont
        \setlength{\tabcolsep}{4pt}
        \resizebox{\linewidth}{!}{
		\begin{tabular}{@{}ccccccccccccccccccccccc@{}}
	  \toprule
        \multicolumn{2}{c}{\multirow{3}{*}{\textbf{Model}}} & \multicolumn{16}{c}{\textbf{Language Pair}} & \multicolumn{5}{c}{\multirow{2}{*}{\textbf{Average Score}}} \\ \cline{3-18}
        \multicolumn{2}{c}{} & \multicolumn{4}{c}{English\(\Rightarrow\)Oromo $^\spadesuit$} & \multicolumn{4}{c}{English\(\Rightarrow\)Welsh} & \multicolumn{4}{c}{English\(\Rightarrow\)Urdu} & \multicolumn{4}{c}{English\(\Rightarrow\)Swahili} & \multicolumn{5}{c}{} \\ \cmidrule{3-23} 
        \multicolumn{2}{c}{} & R-1 & R-2 & R-L & BS & R-1 & R-2 & R-L & BS & R-1 & R-2 & R-L & BS & R-1 & R-2 & R-L & BS & R-1 & R-2 & R-L & S-R & BS \\ \midrule \midrule
        \multirow{2}{*}{mBART-50} & 0-shot & 1.90 & 0.24 & 1.22 & 55.16 & 1.77 & 0.08 & 1.77 & 60.84 & 0.09 & 0.00 & 0.09 & 60.98 & 1.65 & 0.20 & 1.65 & 59.82 & 1.35 & 0.13 & 1.18 & 2.66 & 59.20 \\ 
         & 1000-shot & 5.66 & 0.56 & 4.85 & 56.48 & 18.38 & 3.60 & 13.90 & 68.68 & 18.09 & 3.25 & 14.18 & 68.47 & 16.89 & 4.80 & 13.41 & 69.55 & 14.76 & 3.05 & 11.59 & 29.40 & 65.80 \\ \midrule
        {mT5-base} & 1000-shot & 0.00 & 0.00 & 0.00 & 48.11 & 6.86 & 0.60 & 6.86 & 60.19 & 5.25 & 0.00 & 4.89 & 59.90 & 0.01 & 0.00 & 0.01 & 47.39 & 3.03 & 0.15 & 2.94 & 6.12 & 53.90 \\ \midrule
        \multirow{3}{*}{\gptthreepointfive \cite{park}} & zero-shot & 3.56 & 0.92 & 3.33 & 55.12 & 9.78 & 2.07 & 6.62 & 65.87 & 14.36 & 3.39 & 9.36 & 66.21 & 14.66 & 4.23 & 9.97 & 67.56 & 10.59 & 2.65 & 7.32 & 20.56 & 63.69 \\ 
         & one-shot & 4.10 & 0.36 & 3.52 & 57.91 & 14.23 & 4.21 & 10.54 & 67.13 & 13.84 & 2.25 & 9.92 & 65.01 & 19.77 & 5.49 & 14.11 & 69.38 & 12.99 & 3.08 & 9.52 & 25.59 & 64.86 \\ 
         & two-shot & 3.02 & 0.89 & 2.97 & 54.22 & 15.19 & 4.18 & 10.87 & 67.09 & 14.91 & 2.79 & 10.41 & 66.12 & 18.06 & 4.97 & 12.79 & 68.87 & 12.80 & 3.21 & 9.26 & 25.27 & 64.08 \\ \midrule
        \multirow{3}{*}{\gptfour \cite{park}} & zero-shot & 5.94 & 1.14 & 4.12 & 61.69 & 9.58 & 2.20 & 6.59 & 65.86 & 13.24 & 3.18 & 8.44 & 66.04 & 9.44 & 2.65 & 6.81 & 66.01 & 9.55 & 2.29 & 6.49 & 18.33 & 64.90 \\ 
         & one-shot & 7.26 & 1.31 & 4.96 & 62.20 & 15.35 & 4.35 & 10.51 & 67.50 & 15.49 & 3.84 & 10.07 & 66.92 & 15.62 & 4.86 & 11.27 & 68.08 & 13.43 & 3.59 & 9.20 & 26.22 & 66.18 \\ 
         & two-shot & 6.89 & 1.04 & 4.59 & 62.38 & 15.00 & \uline{4.61} & 10.11 & 67.60 & 16.19 & 3.76 & 11.02 & 67.32 & 17.33 & 5.06 & 11.97 & 68.58 & 13.85 & 3.62 & 9.42 & 26.89 & 66.47 \\ \midrule
         {\gptthreepointfive w/ summarize-translate} & zero-shot & 4.20 & 1.14 & 3.13 & 57.31 & 6.05 & 0.84 & 4.52 & 63.02 & 18.60 & 3.38 & 12.91 & 68.12 & 16.60 & 4.72 & 11.52 & 67.77 & 11.36 & 2.52 & 8.02 & 21.90 & 64.06 \\ 
        {\gptfour w/ summarize-translate} & zero-shot & 4.52 & 0.86 & 3.43 & 61.37 & 10.77 & 2.51 & 7.05 & 66.04 & 12.50 & 2.83 & 8.55 & 66.42 & 11.71 & 3.33 & 8.73 & 66.78 & 9.88 & 2.38 & 6.94 & 19.20 & 65.15 \\ \midrule
         {\gptthreepointfive w/ \sitr (Ours)} & zero-shot & \uline{13.25} & \textbf{3.92} & \textbf{10.35} & \uline{66.69} & \textbf{26.11} & \textbf{4.96} & \textbf{16.83} & \textbf{70.42} & \uline{22.51} & \uline{3.94} & \uline{14.49} & \uline{69.20} & \uline{23.21} & \uline{5.84} & \uline{15.95} & \uline{70.03} & \uline{21.27} & \textbf{4.67} & \uline{14.41} & \uline{40.35} & \uline{69.09} \\ 
        {\gptfour w/ \sitr (Ours)} & zero-shot & \textbf{15.74} & \uline{3.33} & \textbf{11.00} & \textbf{67.50} & \uline{23.24} & 4.40 & \uline{15.17} & \uline{69.41} & \textbf{23.09} & \textbf{4.53} & \textbf{15.47} & \textbf{69.29} & \textbf{23.91} & \textbf{6.01} & \textbf{16.63} & \textbf{70.39} & \textbf{21.50} & \uline{4.57} & \textbf{14.57} & \textbf{40.64} & \textbf{69.15} \\ \bottomrule
		\end{tabular}
        \vspace{-2mm}
        }
	\caption{Additional experimental results on the CrossSum dataset. R-1, R-2, R-L, S-R and BS refer to ROUGE-1, ROUGE-2, ROUGE-L, sum of ROUGE-1/2/L and BERTScore respectively. The task with $^\spadesuit$ means training data less than 1000, where 1000-shot setting equals full fine-tuning. The best result on every target language is highlighted in \textbf{bold} font, and the second best result is marked with an \uline{underline}.}\label{tab:crosssumappendixtwo}
\end{table*}

\section{Prompt vs Meta-Generation}\label{moreablation}
To further explore the impact of effective prompt guidance, we conduct three sets of comparative experiments for comprehensive ablation studies:

(i) Replace the designed summarization prompt with a simple summarization prompt (summarize the following text ...). 

(ii) Replace the designed translation prompt with a simple translation prompt (translate the following text into ...). 

(iii) Replace both summarization and translation prompts with their respective simple versions.

\begin{table*}[!tb]
	\centering
	\begin{adjustbox}{width=0.75\linewidth}
    \renewcommand{\arraystretch}{1.15}
        \scriptsize
        \setlength{\tabcolsep}{0pt}
        \resizebox{\linewidth}{!}{
    		\begin{tabular}{@{}cccccc@{}}
    		\toprule
            \multirow{3}{*}{\textbf{Model}} & \multirow{3}{*}{\textbf{Method}} & \multicolumn{4}{c}{\textbf{Dataset}} \\ \cmidrule{3-6}
             &  & \multicolumn{2}{c}{\textbf{CrossSum}} & \multicolumn{2}{c}{\textbf{WikiLingua}} \\ \cmidrule{3-6} 
             &  & {\,\,\,\,SUM-ROUGE\,\,\,\,} & {\,\,\,\,BERTScore\,\,\,\,} & {\,\,\,\,SUM-ROUGE\,\,\,\,} & {\,\,\,\,BERTScore\,\,\,\,} \\ \midrule \midrule
            \multirow{4}{*}{\,\,\,\,\gptthreepointfive\,\,\,\,} & \sitr & \textbf{32.15} & \textbf{69.58} & \textbf{49.02} & \textbf{71.45} \\
             & \cona & 30.17 & 69.01 & 46.71 & 70.52 \\
             & \conb & 29.91 & 68.83 & 45.10 & 70.21 \\
             & \conc & 27.45 & 68.76 & 43.29 & 70.03 \\ \midrule
            \multirow{4}{*}{\,\,\,\,\gptfour\,\,\,\,} & \sitr & \textbf{34.54} & \textbf{70.13} & \textbf{45.38} & \textbf{70.53} \\
             & \cona & 31.23 & 69.71 & 41.83 & 69.92 \\
             & \conb & 32.09 & 69.59 & 42.37 & 69.88 \\
             & \conc & 29.85 & 69.26 & 39.13 & 69.36 \\ \bottomrule
    		\end{tabular}
            \vspace{-2mm}
        }
	\end{adjustbox}
	\caption{Experimental results on two datasets for prompt replacement. SUM-ROUGE refers to the sum of ROUGE-1/2/L scores. The best result is highlighted in \textbf{bold} font.\label{tab:promptablationcombine}}
\end{table*}

The ablation results for prompt replacements are shown in Table \ref{tab:promptablationcombine}, comparing the performance of the original {\sitr} method with the three prompt variations. From these results, we can draw the following conclusions:

(1) Using appropriate prompts for both the \textbf{{\summarization}} and \textbf{{\translation}} steps positively impacts the large language model's performance in cross-lingual summarization tasks for low-resource languages.

(2) The summarization prompt more significantly affects the quality of the final output than the translation prompt, given its role as the initial step in the architecture.

(3) Considering the results in Figure \ref{fig:ablationbertscore} and Figure \ref{fig:ablationsumrouge}, we could find that the decrease in performance from prompt replacement is much smaller than the decrease from removing key meta-generation steps. 

In summary, manually designed prompts, along with the \textbf{{\improvement}} and \textbf{{\refinement}} steps, significantly enhance the performance of large language models in cross-lingual summarization tasks for low-resource languages. However, the two meta-generation steps contribute more significantly to improving model capabilities than the prompts themselves.


\begin{figure*}[!ht]
\begin{userquery}
\textbf{Text Generation Prompt for Summarization Step}: 
    
You are tasked with creating a concise summary of a given text. The text to be summarized is provided below: 

\{\{TEXT\_TO\_SUMMARIZE\}\}

To create an effective summary, follow these guidelines:

1. Read the entire text carefully to understand the main ideas and overall message.

2. Identify the key points, main arguments, or central themes of the text.

3. Focus on the most important information and avoid including minor details or examples.

4. Aim to capture the essence of the text in a concise manner.

5. The summary should be significantly shorter than the original text, ideally about 5-10\% of its length.

6. Ensure that the summary flows logically and maintains coherence.

7. Do not include your own opinions or interpretations; stick to the information presented in the original text.

Write your summary within <summary> tags. The summary should be brief and to the point, covering only the main content without delving into excessive details. Aim for as few sentences as possible.
\end{userquery}
\caption{Text Generation Prompt for Summarization Step (Our {\sitr}).}
\label{fig: summarization_prompt}
\end{figure*}

\begin{figure*}[!ht]
\begin{userquery}
\textbf{Text Generation Prompt for Improvement Step}: 

You are tasked with evaluating and improving a summary of a given text. Your goal is to create a brief, concise summary that captures the main points without unnecessary details. Follow these steps:

1. First, read the original text: 

\{\{SOURCE\_TEXT\}\}

2. Now, read the current summary: 

\{\{SUMMARY\}\}

3. Evaluate the current summary based on the following criteria:

   \quad a. Accuracy: Does it correctly represent the main ideas of the original text?
   
   \quad b. Conciseness: Is it brief and to the point?
   
   \quad c. Clarity: Is it easy to understand?

4. Improve the summary by:

   \quad a. Removing any unnecessary details or redundant information
   
   \quad b. Limiting the length to an equal number or fewer sentences

5. Provide your improved summary within \textless improved\_summary\textgreater{} tags.

Remember, the goal is to create a brief and accurate summary that captures the essence of the original text without going into details.

\end{userquery}
\caption{Text Generation Prompt for Improvement Step (Our {\sitr}).}
\label{fig: improvement_prompt}
\end{figure*}

\begin{figure*}[!ht]
\begin{userquery}
\textbf{Text Generation Prompt for Translation Step}: 

You are a highly skilled translator with expertise in various languages, including less commonly used ones. Your task is to translate an English text into a specified target language. Please follow these instructions carefully:

1. You will be provided with an English text to translate. The text is as follows: 

\{\{SOURCE\_TEXT\}\}

2. The target language for translation is: 

\{\{TARGET\_LANGUAGE\}\}

3. When translating, please consider the following:

   \quad - Pay attention to cultural nuances and idiomatic expressions
   
   \quad - Maintain the original tone and style of the text as much as possible
   
   \quad - Ensure grammatical accuracy in the target language
   
   \quad - If there are any terms or concepts that don't have a direct equivalent in the target language, provide the best possible translation and include a brief explanation in parentheses

4. Your output should adhere to these guidelines:

   \quad - Do not repeat words or sentences unnecessarily
   
   \quad - Avoid any gibberish or nonsensical text
   
   \quad - Provide a fluent and coherent translation
   
   \quad - If you're unsure about a particular word or phrase, provide your best translation and indicate your uncertainty with [?] after the word or phrase

5. Please provide your translation within <translation> tags. If you need to include any translator's notes or explanations, please add them after the translation within <notes> tags.

Now, please translate the given English text into the specified target language.

\end{userquery}
\caption{Text Generation Prompt for Translation Step (Our {\sitr}).}
\label{fig: translation_prompt}
\end{figure*}

\begin{figure*}[!ht]
\begin{userquery}
\textbf{Text Generation Prompt for Refinement Step}: 

You are a bilingual expert in English and \{\{TARGET\_LANGUAGE\}\}. Your task is to analyze and refine a translation from English to \{\{TARGET\_LANGUAGE\}\}, focusing on fixing any duplicate content and gibberish. Follow these steps:

1. First, carefully read the original English text: 

\{\{ENGLISH\_TEXT\}\}

2. Now, examine the translation in \{\{TARGET\_LANGUAGE\}\}: 

\{\{TRANSLATED\_TEXT\}\}

3. Analyze the translation for the following issues:

   \quad a. Overall accuracy: Check if the translation accurately conveys the meaning of the original English text.
   
   \quad b. Gibberish: Look for any parts of the translation that don't make sense or seem like nonsensical text.

4. Refine the translation by:

   \quad a. Making minor adjustments to improve accuracy and fluency, while preserving the original style and tone
   
   \quad b. Replacing gibberish with appropriate \{\{TARGET\_LANGUAGE\}\} text that matches the meaning of the English original

5. Provide your refined translation inside \textless refined\_translation\textgreater{} tags.

\end{userquery}
\caption{Text Generation Prompt for Refinement Step (Our {\sitr}).}
\label{fig: refinement_prompt}
\end{figure*}

\begin{figure*}[!ht]
\begin{userquery}
\textbf{Text Generation Prompt for Summarize-Translate Method}: 

Please first summarize the following text and then translate the summary into \{\{TARGET\_LANGUAGE\}\}: 

\{\{TEXT\_TO\_SUMMARIZE\}\}. 

Return the final translated \{\{TARGET\_LANGUAGE\}\} summary within \textless translated\_summary\textgreater{} tags.

\end{userquery}
\caption{Text Generation Prompt for Single-Step Summarize-Translate Method.}
\label{fig: st_prompt}
\end{figure*}

\begin{figure*}[!ht]
\begin{userquery}
\textbf{Text Generation Prompt for Few-Shot Method}: 

Please summarize the following text in \{\{TARGET\_LANGUAGE\}\}.

Example 1

Text: \{\{EXAMPLE1\_TEXT\}\}

Translated summary: \{\{EXAMPLE1\_SUMMARY\}\}

Example 2

Text: \{\{EXAMPLE2\_TEXT\}\}

Translated summary: \{\{EXAMPLE2\_SUMMARY\}\}

Test Text

Text: \{\{TEST\_TEXT\}\}

Translated summary:
\end{userquery}
\caption{Text Generation Prompt for Few-Shot Method.}
\label{fig: fewshot_prompt}
\end{figure*}

\begin{figure*}[!tb]
	\centering
	\includegraphics[width=1\linewidth]{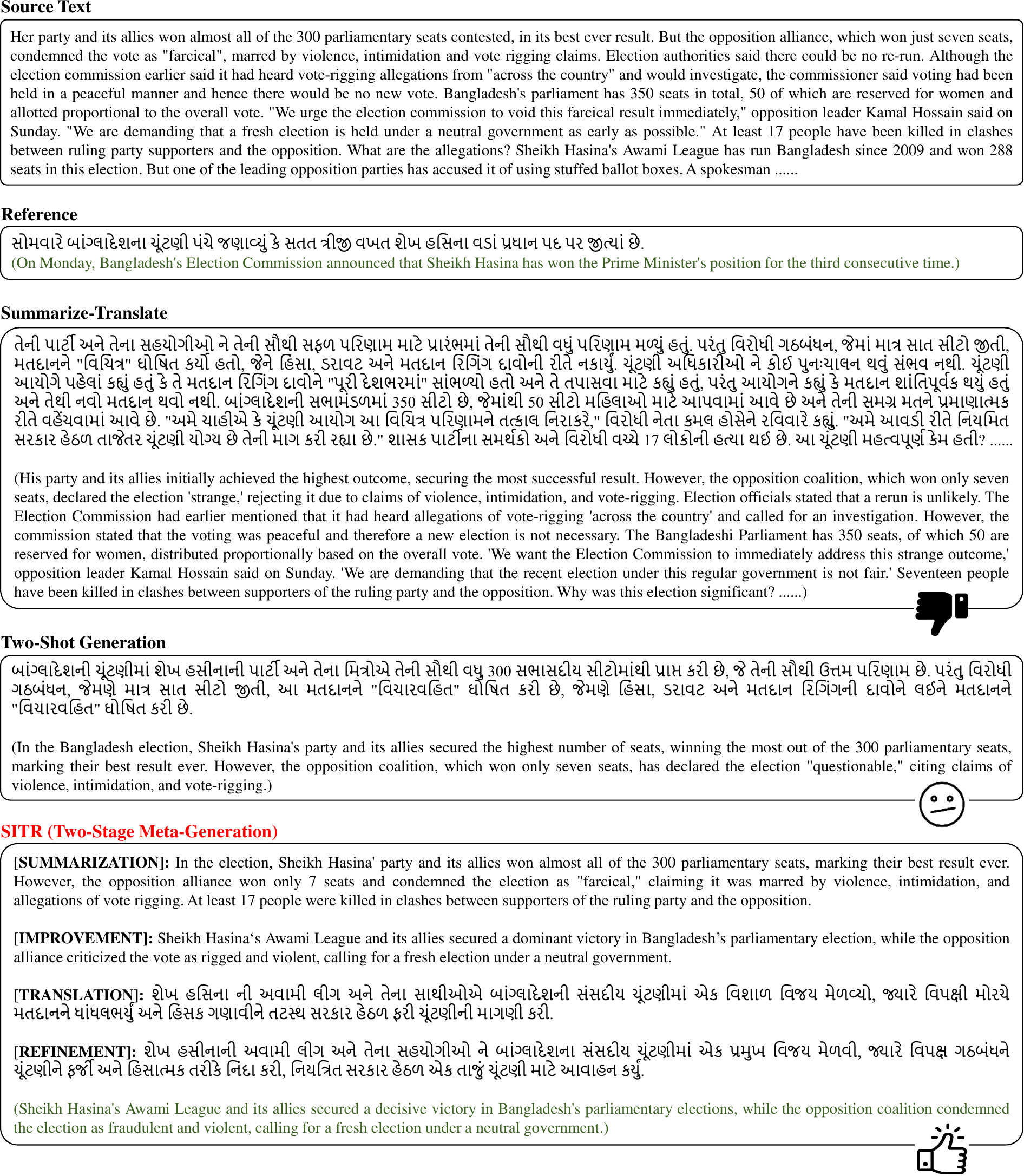}
	\caption{Comparison of three different LLM methods on one single test example to summarize English source text in Gujarati. Our \sitr method performs the best, the result is even better than the reference. The English translation of each model output is shown in brackets.
    \vspace{-2mm}
	\label{fig:appendixexample}}
\end{figure*}

\end{document}